\documentclass[10pt,twocolumn,letterpaper]{article}
\usepackage[pagenumbers]{cvpr}

\usepackage{cuted} %

\usepackage{currfile} %

\usepackage{caption} %

\usepackage{array}
\usepackage{tabularx}
\usepackage{multirow}
\usepackage{makecell}
\usepackage{colortbl}
\usepackage{algorithm}
\usepackage{pifont}
\newcolumntype{Y}{>{\centering\arraybackslash}X}
\definecolor{solblue}{HTML}{E5EFF9}
\definecolor{solgray}{gray}{0.92}
\newcommand{\WCNumber}[1]{\raisebox{-0.5pt}{\scalebox{1.1}{\ding{\numexpr181+#1\relax}}}}
\newsavebox{\soltablebox}
\newenvironment{soltabularfit}{%
  \begin{lrbox}{\soltablebox}%
}{%
  \end{lrbox}%
  \ifdim\wd\soltablebox>\linewidth
    \resizebox{\linewidth}{!}{\usebox{\soltablebox}}%
  \else
    \usebox{\soltablebox}%
  \fi
}

\usepackage{bbding}
\definecolor{cvprblue}{rgb}{0.21,0.49,0.74}
\usepackage[pagebackref,breaklinks,colorlinks,allcolors=cvprblue]{hyperref}
\title{Sol\textsuperscript{3}: Accelerating Video Diffusion for Free via Hierarchical Sparse Attention Routing}
\author{%
  \raisebox{-5pt}[0pt][0pt]{%
    \footnotesize
    \begin{tabular}{@{}c@{}}
      {\large Guoxin Fan\textsuperscript{1,2,*}\quad
       Weilun Feng\textsuperscript{1,2,*}\quad
       Chuanguang Yang\textsuperscript{1,\textdagger}\quad
       Mingqiang Wu\textsuperscript{1,2}\quad
       Xiaokun Liu\textsuperscript{1,2}}\\
      {\large Jitao Lu\textsuperscript{1,2}\quad
       Zhulin An\textsuperscript{1}\quad
       Libo Huang\textsuperscript{1}\quad
       Boyu Diao\textsuperscript{1}\quad
       Di Gao\textsuperscript{3}\quad
       Yongjun Xu\textsuperscript{1,4}}\\[2pt]
      \textsuperscript{1}State Key Laboratory of AI Safety, Institute of Computing Technology, Chinese Academy of Sciences\quad
      \textsuperscript{2}University of Chinese Academy of Sciences\\
      \textsuperscript{3}Hangzhou Institute for Advanced Study, University of the Chinese Academy of Sciences\quad
      \textsuperscript{4}Xiamen Institute of Data Intelligence, Xiamen, China\\[2pt]
      \textsuperscript{*}Equal contribution.\quad
      \textsuperscript{\textdagger}Corresponding author: \href{mailto:yangchuanguang@ict.ac.cn}{yangchuanguang@ict.ac.cn}.
    \end{tabular}%
  }%
}
\hypersetup{%
  pdftitle={Sol³: Accelerating Video Diffusion for Free via Hierarchical Sparse Attention Routing},
  pdfauthor={Guoxin Fan, Weilun Feng, Chuanguang Yang, Mingqiang Wu, Xiaokun Liu, Jitao Lu, Zhulin An, Libo Huang, Boyu Diao, Di Gao, Yongjun Xu},
  pdfsubject={},
  pdfkeywords={}
}
\begin{document}
\maketitle
\begin{strip}
\centering
\includegraphics[width=\linewidth]{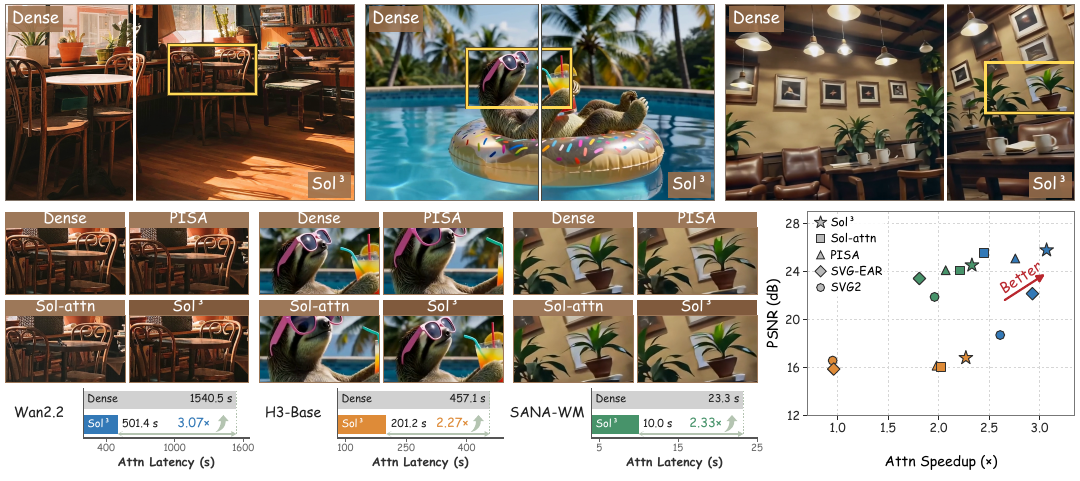}
\captionsetup{hypcap=false}
\captionof{figure}{\textbf{Sol\textsuperscript{3}} speeds up diffusion models, including video generation, audio-video generation, and world simulation, achieving up to \textbf{3.07\texttimes{}} faster attention processing while preserving high-fidelity details.}
\label{fig:teaser}
\captionsetup{hypcap=true}
\end{strip}
\begin{abstract}
Sol-Attn (Sparsifying online attention) combines sparse exact computation with low-cost representative approximation to accelerate training-free inference in diffusion video models. However, intra-block attention can focus on only a few tokens. While preserving these key interactions exactly, the block-level decision incurs the cost of evaluating many low-response members. We propose \textbf{Sol\textsuperscript{3}}, a training-free hierarchical sparse attention method that further refines how exact computations are allocated. First, we introduce \textbf{Gaussian-guided hierarchical routing}, which determines the scope of exact computation step by step, from blocks to tokens, using statistical thresholds at each level. Only a subset of members within Mixed blocks that require refinement is selected for exact computation, while representative terms approximate the remaining members. We further design \textbf{heterogeneous routing statistics} that combine block-center correlations with variations in Key and Value within blocks to identify blocks that are difficult to approximate. Among these Mixed blocks, we uniformly compare raw token relevance to select members requiring exact computation. Experiments on video generation, audio-video co-generation, and world simulation tasks demonstrate that \textbf{Sol\textsuperscript{3}} achieves \textbf{86\%} attention sparsity and delivers a \textbf{3.07\texttimes{}} attention speedup while preserving quality.
Our code is available at \href{https://github.com/FofGofx/sol3}{https://github.com/FofGofx/sol3}.
\end{abstract}
\vspace{-12pt}

\section{Introduction}
\label{sec:source6}
Training-free sparse attention accelerates inference by selectively computing attention interactions without updating pre-trained model parameters. Sol-Attn~\cite{solattn} (Sparsifying online attention) is a promising method: it dynamically sets thresholds from the statistical distribution of block-center correlations, then combines sparse exact computation with low-cost approximation in a jointly normalized online computation. This reduces computational cost while preserving contributions from regions skipped by exact computation. The method has been validated across various diffusion models, including Sol-H3~\cite{solh3}.

However, since Sol-Attn makes a unified decision between block-level exact computation and block-level approximation, it struggles to adapt to the varying computational requirements of intra-block interactions. To further refine the allocation of computations, the following two interrelated challenges must be addressed:

\par\noindent \textbf{\WCNumber{1}. A few dominant features within and between blocks.}
Video attention varies with how queries align with spatiotemporal content, and its response distribution does not always align with computational block boundaries. Attention imbalance appears both across and within blocks: a small number of blocks account for most attention, while some blocks with overall modest responses still contain a few highly weighted tokens. This distribution creates granularity constraints on block-level decisions: exact block-level computation requires many low-response interactions, while block-level approximation fails to preserve the exact contributions of individual local interactions.

\par\noindent \textbf{\WCNumber{2}. Block-center correlations cannot directly guide fine-grained computational allocation.}
While blocks may have similar overall responses, interactions within them may not be equally easy to approximate. Block-center correlation does not fully account for differences in keys and values within a block. Therefore, calculating allocations solely from overall responses may still leave hard-to-approximate interactions to representative items. Among these blocks that are difficult to approximate, the number of high-response tokens also varies. Within the same query group, each block has a different response baseline and fluctuation range; therefore, each block's threshold reflects only that block's score distribution. Consequently, independent block-by-block filtering may over-allocate members to blocks with weaker responses, while leaving some strong interactions in other blocks to be approximated by representative items.

To address these challenges, we propose \textbf{Sol\textsuperscript{3}}, a training-free hierarchical sparse attention framework for accelerating diffusion video models. Its core principle is to refine computational granularity on demand, based on interaction differences, and to select suitable statistical criteria at each level according to the evaluation objectives. To address intra-block interaction differences, we introduce \textit{Gaussian-Guided Hierarchical Routing} (GHR), which constructs Gaussian thresholds from the distribution of candidate scores at each level. GHR first selects Full blocks that require exact computation for the entire block, then screens the remaining blocks for Mixed blocks, and finally selects the tokens within these blocks that require exact computation. We further designed \textit{Heterogeneous Routing Statistics} (HRS) to construct criteria aligned with the computational objectives at each level: we measure overall responsiveness using block-center correlation. We assess representative approximation difficulty by combining block proxy weights, Value dispersion, and Key variance in the query direction. Finally, we uniformly compare raw token relevance across Mixed blocks under a common query reference.

\noindent Our contributions can be summarized as follows:
\begin{itemize}[leftmargin=\leftmargini]
\item We point out that block-level decision-making cannot fully leverage the few dominant features within a block, and we identify differences in the information used by various decision objectives during refinement.
\item We propose a Gaussian-guided hierarchical routing approach that refines the calculation scope on demand from the block to the token level, enabling exact computation and representative approximations to be allocated collaboratively within a block.
\item We propose heterogeneous routing statistics that combine key and value variations within blocks to estimate representative approximation difficulty, and select members for exact computation across Mixed blocks based on raw token relevance to the query group.
\item Extensive experiments across various diffusion video models demonstrate that Sol\textsuperscript{3} can significantly reduce the computational cost of attention while maintaining video-generation quality.
\end{itemize}

\begin{figure*}[t!]
\centering
\includegraphics[width=\linewidth]{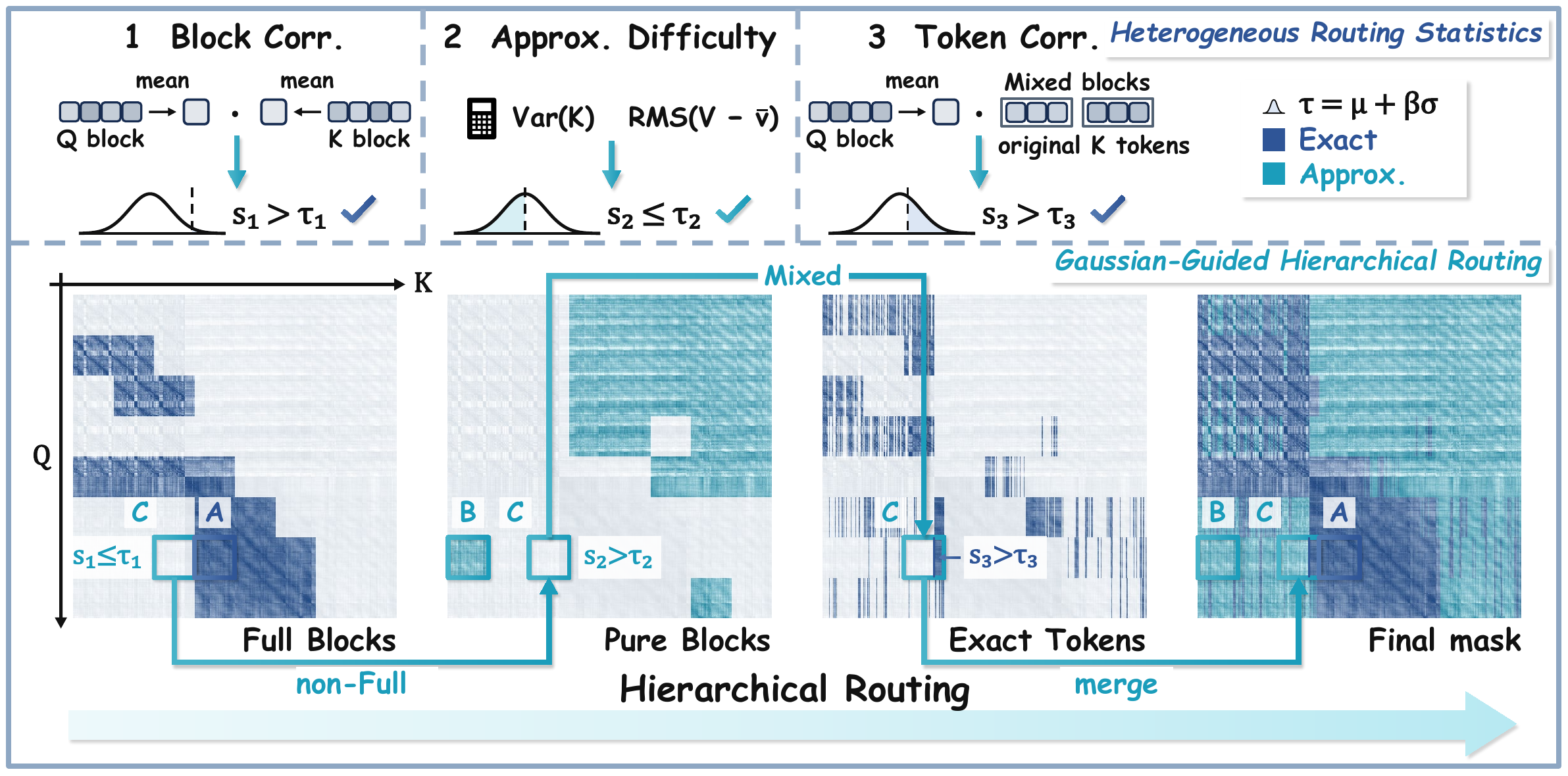}
\caption{\textbf{Overview of the Sol\textsuperscript{3} framework.} \textbf{(Top)} \textbf{Heterogeneous Routing Statistics} (HRS) computes block-center correlation from block means, combines it with key variance in the query direction and within-block value dispersion, and uses these signals to assess representative approximation difficulty while comparing raw token relevance across Mixed blocks. \textbf{(Bottom)} \textbf{Gaussian-Guided Hierarchical Routing} (GHR) selects Full blocks in sequence and filters Mixed blocks using scores at each level and statistical thresholds; it then selects tokens from the Mixed blocks for exact computation while approximating the remaining interactions with representative items.}
\label{fig:method}
\end{figure*}
\section{Related Work}
\label{sec:source27}
\subsection{Sparse and Hybrid Attention for Video Diffusion}
\label{sec:source28}
Sparse attention in video diffusion selects regions for exact computation using structural priors~\cite{sta,sparsevdit} or input content~\cite{adaspa,vsa,vmoba,dsv}. SVG~\cite{svg} selects sparse spatiotemporal patterns online. Radial Attention~\cite{radial} builds static interaction ranges from spatiotemporal decay priors. SVG2~\cite{svg2}, SVOO~\cite{svoo}, and AdaCluster~\cite{adacluster} use content-based clustering and rearrangement to organize regions for exact computation.

Hybrid attention further compensates for unaccounted interactions: SLA~\cite{sla} uses exact computation, linear compensation, and path skipping; SLA2~\cite{sla2} learns routing and the mixing ratio of learning branches; PISA~\cite{pisa} combines exact computation with Taylor approximation under joint normalization; SVG-EAR~\cite{svgear} compensates for unaccounted regions using cluster centers; and Sol-Attn~\cite{solattn} integrates routing, exact computation, and approximation into an online softmax~\cite{onlinesoftmax}.

However, attention computation does not always align with block boundaries, making it difficult to balance efficiency and information preservation with block-wise exact computation or approximation. This has prompted us to explore fine-grained computational allocation that can adapt to intra-block heterogeneity.

\subsection{Routing Criteria for Sparse Attention}
\label{sec:source32}
Routing performance depends on proxy structure~\cite{minference,quest} and on the scoring objective and selection rules~\cite{flexprefill,tactic,twilight}. SpargeAttn~\cite{sparge} uses block-wise averaging to estimate attention and identifies blocks unsuitable for compression based on self-similarity. XAttention~\cite{xattention} uses anti-diagonal sampling. SVG2~\cite{svg2}, SVOO~\cite{svoo}, and AdaCluster~\cite{adacluster} build proxies through central interaction, QK collaborative clustering, and query/key-differentiated clustering, respectively.

Regarding scoring objectives and selection rules, SVG-EAR~\cite{svgear} combines estimated error and computational cost to allocate exact computation; SLA2~\cite{sla2} learns to allocate sparse and linear branches; SpargeAttention2~\cite{sparge2} takes the union of blocks selected by Top-k and Top-p; and Sol-Attn~\cite{solattn} constructs dynamic thresholds from query-related distribution statistics to support streaming selection.

Existing methods improve routing through correlation estimation and error awareness. However, as computational allocation becomes more refined, overall importance, approximation reliability, and local interaction requirements reflect different kinds of information. With this in mind, Sol\textsuperscript{3} aligns routing decisions with their underlying criteria.

\section{\texorpdfstring{Sol\textsuperscript{3}}{Sol3}}
\label{sec:source60}
\subsection{Gaussian-Guided Hierarchical Routing (GHR)}
\label{sec:source63}
\par
\begin{figure}[!htbp]
\centering
\includegraphics[width=\linewidth]{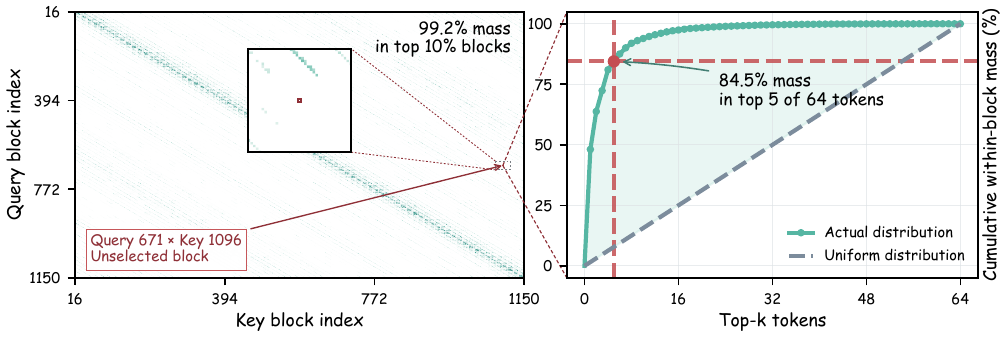}
\caption{\textbf{Illustration of a few dominant features across and within blocks.} In the left figure, attention focuses on a few blocks; the right figure shows a close-up of a block marked in the left figure, where the overall response is subtle, yet attention within that block remains focused on a few tokens.}
\label{fig:o1}
\end{figure}

\noindent \textbf{Observation \WCNumber{1}. \textit{Attention exhibits a distribution characterized by a small number of dominant patterns both across blocks and within blocks.}}
As shown in Figure \ref{fig:o1}, a small number of computation blocks account for much of the attention mass, and even in blocks where the overall response is not especially strong, that mass remains concentrated in a few tokens. For these blocks, block-level exact computation would process many low-response interactions at once, while block-level approximation cannot distinguish or handle local high-response interactions. This led us to adopt a hierarchical routing approach from blocks to tokens, refining the scope of exact computation as needed.

\par\noindent \textbf{Gaussian-Guided Threshold}
To accommodate the varying meanings and numerical scales of scores across levels, we select portions that warrant exact computation or further inspection based on each score's position within the current candidate distribution.
For query group $i$, let $\mathcal C_i^{(\ell)}$ be the $\ell$th-level candidate set, and $s_{iu}^{(\ell)}$ be the score of candidate $u$. The mean and standard deviation of the candidate scores are denoted by $\mu_i^{(\ell)}$ and $\sigma_i^{(\ell)}$, and the threshold and selected set are defined as:

\begin{equation}
\begin{gathered}
\tau_i^{(\ell)}=\mu_i^{(\ell)}+\beta_\ell\sigma_i^{(\ell)},\\[6pt]
\mathcal S_i^{(\ell)}
=\{u\in\mathcal C_i^{(\ell)}:s_{iu}^{(\ell)}>\tau_i^{(\ell)}\},
\end{gathered}
\label{eq:source75}
\end{equation}
where $\beta_{\ell}$  is the normalized offset parameter that adjusts the screening strength. When the normalized distribution of candidate scores approximates a Gaussian distribution, $\beta_{\ell}$ can be set based on the Gaussian reference selection rate $\rho_{\mathrm{G},\ell}$ over the current candidate set:

\begin{equation}
\rho_{\mathrm G,\ell}=1-\Phi(\beta_\ell),
\qquad
\beta_\ell=\Phi^{-1}(1-\rho_{\mathrm G,\ell}),
\label{eq:source77}
\end{equation}
where $\Phi$ is the cumulative distribution function of the standard normal distribution.

\par\noindent \textbf{Three-Level Routing from Blocks to Tokens}
For query group i, the first level selects Full blocks $\mathcal{F}_{i}$ whose block-center correlation exceeds the threshold and includes all members in exact computation. The second level filters Mixed blocks $\mathcal{M}_{i}$ whose representative approximation difficulty exceeds the threshold from the remaining blocks, while approximating the remaining Pure blocks using their representatives. Let $\mathcal{B}_{j}$ be the set of token indices for the jth key-value block. At the third level, members whose raw token relevance exceeds the threshold are filtered from the unified candidate set $\mathcal C_i^{(3)}=\bigcup_{j\in\mathcal M_i}\mathcal B_j$ to form $\mathcal{S}_{i}^{(3)}$, while representative items approximate the remaining members. The final set of tokens for exact computation is:

\begin{equation}
\mathcal E_i=
\left(\bigcup_{j\in\mathcal F_i}\mathcal B_j\right)
\cup\mathcal S_i^{(3)}.
\label{eq:source84}
\end{equation}
\par\noindent \textbf{Compact Implementation of Hierarchical Selection}
When multiple Mixed blocks retain only a few members, organizing computations by the original blocks can fragment execution and reduce the benefits of sparse computation. To address this, Sol\textsuperscript{3} organizes the selected members into a compact index for batched exact computation across the original blocks. For details on the index construction and kernel implementation, \textbf{see Appendix \ref{sec:source277}.}

\subsection{Heterogeneous Routing Statistics (HRS)}
\label{sec:source88}
\begin{figure}[!htbp]
\centering
\includegraphics[width=\linewidth]{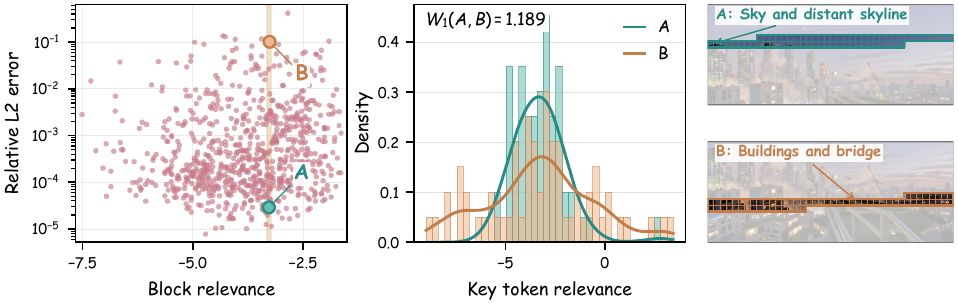}
\caption{\textbf{Block-center correlations and representative approximation errors.} \textbf{(Left)} A and B have similar correlations but different errors. \textbf{(Middle)} We use the first-order Wasserstein distance $W_{1}$ to measure differences between blocks in the distribution of raw token relevance. \textbf{(Right)} Token regions for A and B.}
\label{fig:pair}
\end{figure}
\par\noindent \textbf{Observation \WCNumber{2}. \textit{Similar block-center correlations correspond to different approximation errors.}}
As shown in Figure \ref{fig:pair}, interactions with similar block-center correlations still produce significantly different attention output errors, even after approximation with representative terms. Further examination of these blocks reveals differences in the distributions of their internal keys and values that block-center correlations do not fully capture. This motivates us to incorporate intra-block distributions when evaluating how difficult representative approximation is.

\par\noindent \textbf{Representative Approximation Difficulty}
To quantify the difficulty of a representative approximation for a block under the current query, we combine the overall response with changes in block-internal features. Block-center correlation reflects the overall response, while the dispersion of the value and the variation in the key along the query direction characterize the differences in content and weight within the representative approximation, respectively. Based on this, we define the representative approximation difficulty score for second-level routing:

\begin{equation}
s_{ij}^{(2)}=a_{ij}+\log\!\left(n_j R_j\right)+\Delta_{ij},
\label{eq:source97}
\end{equation}
where $\bar{\mathbf q}_i$ and $\bar{\mathbf k}_j\in\mathbb R^d$ represent the mean of the query group Query and the mean of the key block Key, respectively, and d is the feature dimension of Query and Key. Their scaled dot product $a_{ij}=\bar{\mathbf q}_i^\top\bar{\mathbf k}_j/\sqrt d$ measures block-center correlation. $n_{j}$ is the number of tokens in the block, $R_{j}$ is the root-mean-square deviation of the Value from the block mean, and $\Delta_{ij}$ is a correction term based on the Key variance in the query direction. \textbf{See Appendices \ref{sec:source212} and \ref{sec:source252} for the calculations and derivations of each term.}

\begin{figure}[!htbp]
\centering
\includegraphics[width=\linewidth]{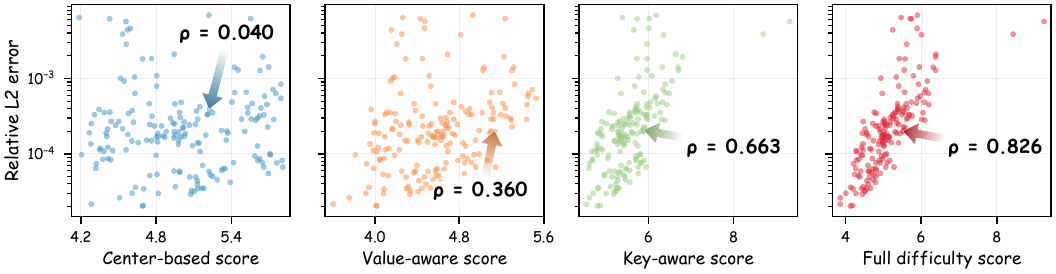}
\caption{\textbf{Visualization of the relationship between representative approximation difficulty and approximation error.} We show the relationship between block-center correlation, value-aware, key-aware, and full-score measures and approximation error, and report the rank correlation coefficient $\rho$.}
\label{fig:proxy}
\end{figure}

As shown in Figure \ref{fig:proxy}, the full representative approximation difficulty score shows a clearer positive relationship with approximation error; its rank correlation with error is higher than that of block-center correlations and variants that use only Value or Key information.

\begin{figure}[!htbp]
\centering
\includegraphics[width=\linewidth]{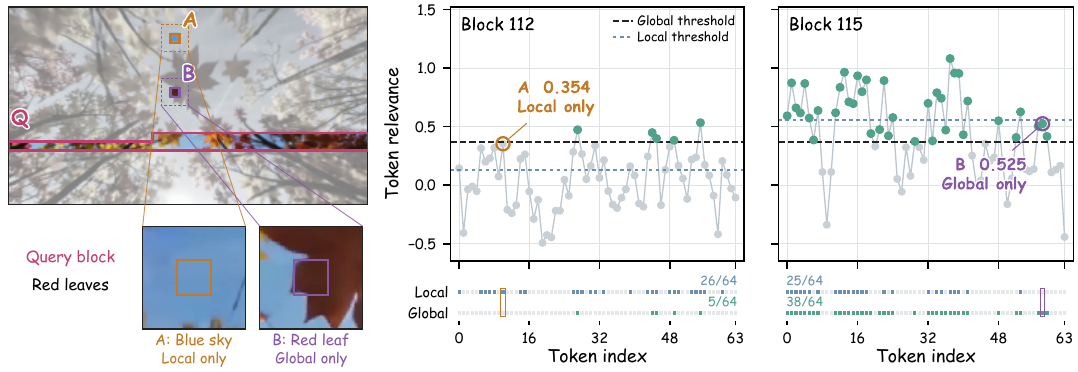}
\caption{\textbf{Illustration of raw token relevance scores and filtering thresholds for key tokens across different Mixed blocks within the same query group.} The left figure shows query group Q with A and B; the middle and right figures show the relevance of the two blocks to Q. Although A is locally prominent, it is weaker than B, which is not locally prominent. Therefore, unified filtering across Mixed blocks retains B rather than A.}
\label{fig:mechanism}
\end{figure}
\par\noindent \textbf{Observation \WCNumber{3}. \textit{The most prominent token within a block does not necessarily have higher raw token relevance.}}
As shown in Figure \ref{fig:mechanism}, for the same query group, Key token A has a high match score in its key-value block, yet its raw token relevance is still lower than that of Key token B, which is not especially prominent in another block. This motivates us to compare raw token relevance across different Mixed blocks under a common query reference.

\begin{figure}[!htbp]
\centering
\includegraphics[width=\linewidth]{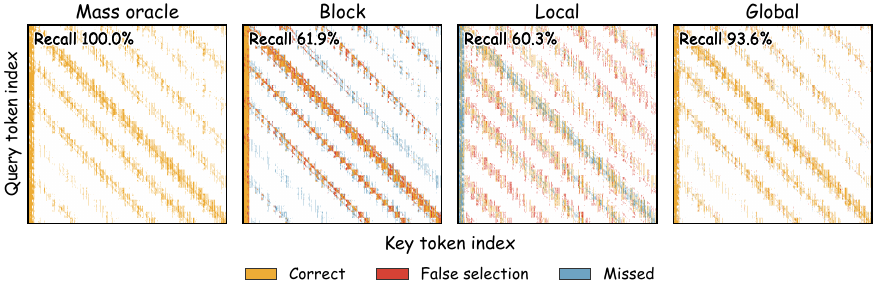}
\caption{\textbf{Visualization of retained key interactions.} Using the Mass oracle as a reference, we show filtering results across the entire block, individual blocks, and uniformly across Mixed blocks.}
\label{fig:selection}
\end{figure}
\par\noindent \textbf{Raw token relevance.} For query group $i$, include all tokens from the Mixed blocks in the same candidate set, and score each token by the scaled dot product of the group's mean $\bar{\mathbf q}_i$ and its raw key $\mathbf k_t$.
\begin{equation}
s_{it}^{(3)}=\frac{\bar{\mathbf q}_i^{\top}\mathbf k_t}{\sqrt d},
\qquad t\in\mathcal C_i^{(3)},
\label{eq:source108}
\end{equation}
where $\mathbf k_t\in\mathbb R^d$ is the original key of token t. The score distribution of the unified candidate set determines the third-level threshold. As shown in Figure \ref{fig:selection}, token-level unified filtering across Mixed blocks preserves key interactions in the reference set more fully than filtering the entire block or filtering each block independently.

\begin{figure}[!htbp]
\centering
\includegraphics[width=\linewidth]{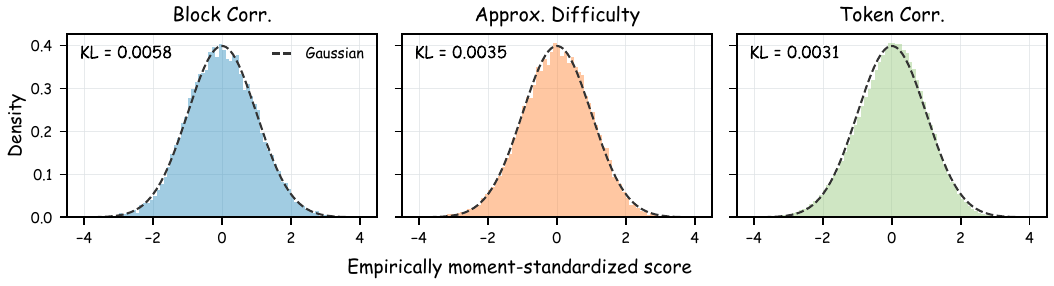}
\caption{\textbf{Visualization of the distribution of three-level routing scores.} We compare the standardized distribution of the three-level scores with a standard Gaussian reference.}
\label{fig:distribution}
\end{figure}
\begin{figure}[!htbp]
\centering
\includegraphics[width=\linewidth]{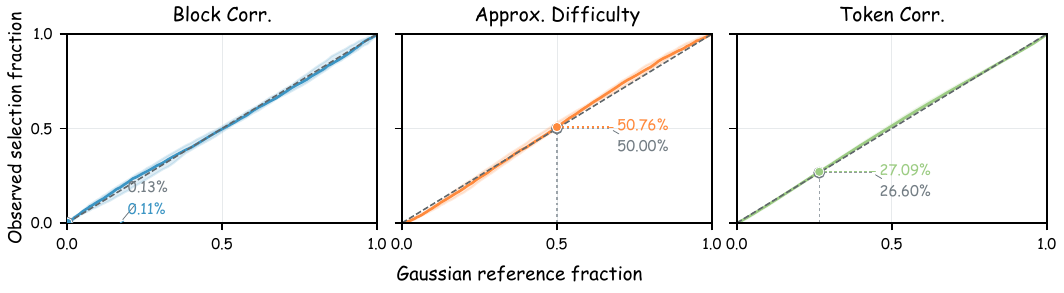}
\caption{\textbf{Calibration of the Gaussian threshold selection ratio.} Compare the Gaussian reference selection ratios at the three routing levels with the actual selection ratios at different thresholds.}
\label{fig:calibration}
\end{figure}
\par\noindent \textbf{Empirical Testing of the Gaussian Threshold}
As shown in Figures \ref{fig:distribution} and \ref{fig:calibration}, the routing scores at the three levels are approximately Gaussian, and the share within the Gaussian reference range is close to the observed share.

\begin{table*}[t!]
\centering
\caption{\textbf{Quantitative comparison of video generation, audio-video generation, and world modeling.} Values outside and inside parentheses report attention and denoising speedups, respectively. Bold and underlined values indicate the best and second-best results, respectively.}
\label{tab:source131}
\begingroup
\small
\setlength{\tabcolsep}{3pt}
\begin{soltabularfit}
\begin{tabular}{l|ccccccccc}
\toprule
\multirow{2}{*}{\textbf{Method}} & \multicolumn{3}{c}{\textbf{Similarity}} & \multicolumn{3}{c}{\textbf{VBench}} & \multicolumn{3}{c}{\textbf{Efficiency}} \\
\cmidrule(lr){2-4}\cmidrule(lr){5-7}\cmidrule(lr){8-10}
 & \textbf{PSNR$\uparrow$} & \textbf{SSIM$\uparrow$} & \textbf{LPIPS$\downarrow$} & \textbf{ImgQual$\uparrow$} & \textbf{SubCons$\uparrow$} & \textbf{AVG$\uparrow$} & \textbf{Sparsity$\uparrow$} & \textbf{FLOPs(P)$\downarrow$} & \textbf{Speedup$\uparrow$} \\
\midrule
\multicolumn{10}{c}{\cellcolor{solgray}Wan2.2 (14B, Image-to-Video, 720p, 81 Frames)} \\
\midrule
FA~\cite{fa2} & $\infty$ & 1.000 & 0.000 & 68.90 & 97.10 & 71.41 & 0.000\% & 521.87 & 1.00$\times$ (1.00$\times$) \\
\midrule
SVG2~\cite{svg2} & 18.72 & 0.478 & 0.549 & 54.44 & 73.36 & 57.70 & 77.49\% & 281.32 & 2.61$\times$ (1.54$\times$) \\
SVG-EAR~\cite{svgear} & 22.17 & 0.647 & 0.270 & 67.31 & 92.94 & 66.23 & 84.13\% & 260.73 & \underline{2.93$\times$} (\underline{1.63$\times$}) \\
PISA~\cite{pisa} & 25.11 & 0.834 & 0.109 & 67.59 & 96.59 & \underline{71.02} & \underline{84.94\%} & \underline{258.19} & 2.76$\times$ (1.51$\times$) \\
Sol-Attn~\cite{solattn} & \underline{25.55} & \underline{0.848} & \underline{0.095} & \textbf{68.81} & \underline{96.84} & 70.93 & 84.57\% & 259.34 & 2.45$\times$ (1.46$\times$) \\
\rowcolor{solblue}
Sol\textsuperscript{3} & \textbf{25.81} & \textbf{0.854} & \textbf{0.086} & \underline{68.31} & \textbf{97.01} & \textbf{71.11} & \textbf{86.37\%} & \textbf{253.77} & {\bfseries\boldmath 3.07$\times$} ({\bfseries\boldmath 1.88$\times$}) \\
\midrule
\multicolumn{10}{c}{\cellcolor{solgray}H3-Base (33B, Text-to-Audio-Video, 768p, 124 Frames)} \\
\midrule
FA~\cite{fa2} & $\infty$ & 1.000 & 0.000 & 73.71 & 91.23 & 74.98 & 0.000\% & 172.34 & 1.00$\times$ (1.00$\times$) \\
\midrule
SVG2~\cite{svg2} & \underline{16.59} & \underline{0.595} & 0.385 & 67.11 & 90.09 & 73.73 & 23.23\% & 152.37 & 0.95$\times$ (0.98$\times$) \\
SVG-EAR~\cite{svgear} & 15.88 & 0.562 & 0.455 & 62.30 & 89.51 & 72.78 & 21.42\% & 153.93 & 0.96$\times$ (0.98$\times$) \\
PISA~\cite{pisa} & 16.17 & 0.571 & \underline{0.371} & 72.99 & 91.08 & 73.91 & \underline{80.27\%} & \underline{102.99} & 1.98$\times$ (\underline{1.32$\times$}) \\
Sol-Attn~\cite{solattn} & 16.06 & 0.570 & 0.384 & \underline{73.08} & \textbf{91.45} & \underline{74.45} & 78.36\% & 104.97 & \underline{2.02$\times$} (\underline{1.32$\times$}) \\
\rowcolor{solblue}
Sol\textsuperscript{3} & \textbf{16.82} & \textbf{0.598} & \textbf{0.348} & \textbf{73.36} & \underline{91.16} & \textbf{74.59} & \textbf{80.78\%} & \textbf{102.59} & {\bfseries\boldmath 2.27$\times$} ({\bfseries\boldmath 1.45$\times$}) \\
\midrule
\multicolumn{10}{c}{\cellcolor{solgray}SANA-WM Refiner (17B, Video-to-Video, 720p, 481 Frames)} \\
\midrule
FA~\cite{fa2} & $\infty$ & 1.000 & 0.000 & 74.53 & 84.54 & 76.82 & 0.000\% & 10.369 & 1.00$\times$ (1.00$\times$) \\
\midrule
SVG2~\cite{svg2} & 21.90 & 0.741 & 0.161 & 74.53 & 84.45 & 76.75 & 71.49\% & 7.2479 & 1.96$\times$ (1.24$\times$) \\
SVG-EAR~\cite{svgear} & 23.44 & 0.795 & 0.123 & 74.85 & 84.65 & 75.90 & 71.60\% & 7.2433 & 1.81$\times$ (1.22$\times$) \\
PISA~\cite{pisa} & \underline{24.13} & 0.824 & 0.105 & \underline{74.93} & 84.69 & 76.82 & \underline{78.02\%} & \underline{6.9562} & 2.07$\times$ (1.26$\times$) \\
Sol-Attn~\cite{solattn} & 24.11 & \underline{0.825} & \underline{0.101} & 74.91 & \textbf{85.05} & \underline{76.95} & 77.75\% & 6.9746 & \underline{2.21$\times$} (\underline{1.28$\times$}) \\
\rowcolor{solblue}
Sol\textsuperscript{3} & \textbf{24.55} & \textbf{0.833} & \textbf{0.097} & \textbf{75.02} & \underline{84.73} & \textbf{77.03} & \textbf{78.19\%} & \textbf{6.9555} & {\bfseries\boldmath 2.33$\times$} ({\bfseries\boldmath 1.36$\times$}) \\
\bottomrule
\end{tabular}
\end{soltabularfit}
\endgroup

\end{table*}
\subsection{Overall Framework}
\label{sec:source118}
\par\noindent \textbf{Pipeline summary.} Sol\textsuperscript{3} computes scores on demand at each level via HRS, and GHR determines the scope of exact computation for each level. The exact terms are then normalized with the representative contributions of the remaining members to produce the attention output. Algorithm \ref{alg:framework} summarizes the procedure.

\begin{algorithm}[t]
\caption{Sol\textsuperscript{3} Framework}
\label{alg:framework}
\small
\begin{tabular}{@{}r@{\enspace}>{\raggedright\arraybackslash}p{\dimexpr\linewidth-2em\relax}@{}}
1: & $\textbf{Input: }Q,K,V;\ b_q,b_{kv};\ \beta_1,\beta_2,\beta_3.$ \\
2: & $\text{Partition }Q,K,V\text{; precompute block statistics.}$ \\
3: & $\textbf{for each }\text{query group }Q_i\textbf{ do}$ \\
4: & $\quad\mathcal C_i^{(1)}\leftarrow\text{all KV block indices}.$ \\
5: & \mbox{$\quad\mathcal F_i\leftarrow
\operatorname{GHR}(\mathcal C_i^{(1)},\mathbf{s}_i^{(1)},\beta_1).$}\hspace{0.87em}\mbox{//\enspace\textcolor[HTML]{808080}{high block-center correlation}} \\
6: & \mbox{$\quad\mathcal C_i^{(2)}\leftarrow\mathcal C_i^{(1)}\setminus\mathcal F_i.$}\hspace{2.27em}\mbox{//\enspace\textcolor[HTML]{808080}{screen by approximation difficulty}} \\
7: & $\quad\mathcal M_i\leftarrow
\operatorname{GHR}(\mathcal C_i^{(2)},\mathbf{s}_i^{(2)},\beta_2)
\ \text{(Eq.~\eqref{eq:source97})}.$ \\
8: & \mbox{$\quad\mathcal C_i^{(3)}\leftarrow
\bigcup\nolimits_{j\in\mathcal M_i}\mathcal B_j.$}\hspace{0.44em}\mbox{//\enspace\textcolor[HTML]{808080}{one candidate set for all Mixed blocks}} \\
9: & $\quad\mathcal S_i^{(3)}\leftarrow
\operatorname{GHR}(\mathcal C_i^{(3)},\mathbf{s}_i^{(3)},\beta_3)
\ \text{(Eq.~\eqref{eq:source108})}.$ \\
10: & $\quad\mathcal E_i\leftarrow
\left(\bigcup\nolimits_{j\in\mathcal F_i}\mathcal B_j\right)\cup\mathcal S_i^{(3)}.$ \\
11: & $\quad r_{ij}\leftarrow n_j-|\mathcal E_i\cap\mathcal B_j|.$ \\
12: & $\quad(N^{\mathrm{ex}},\mathbf Z^{\mathrm{ex}})
\leftarrow\operatorname{Exact}(Q_i,K,V,\mathcal E_i).$ \\
13: & $\quad(N^{\mathrm{rep}},\mathbf Z^{\mathrm{rep}})
\leftarrow\operatorname{Rep}(Q_i,\mathbf r_i).$ \\
14: & $\quad\widehat O_i\leftarrow
\frac{N^{\mathrm{ex}}+N^{\mathrm{rep}}}{\mathbf Z^{\mathrm{ex}}+\mathbf Z^{\mathrm{rep}}}.$ \\
15: & $\textbf{end for}$ \\
16: & $\textbf{return }\operatorname{Stack}_i(\widehat O_i)
\text{ in query order.}$ \\
\end{tabular}
\end{algorithm}

\section{Experiments}
\label{sec:source122}
\subsection{Experimental Setup}
\label{sec:source123}
\par\noindent \textbf{Models.} We evaluate a wide range of state-of-the-art diffusion video models. Video generation includes HunyuanVideo-13B~\cite{hunyuanvideo}, which uses the MM-DiT architecture, and Wan2.2-14B~\cite{wan22}, which uses the Cross-DiT architecture. Audio-video generation includes the single-stream models H3-Base~\cite{h3} and FastH3~\cite{fasth3}, as well as the dual-stream model LTX-2.5-22B~\cite{ltx2}. World simulation uses the camera-controlled model SANA-WM~\cite{sanawm}.

\par\noindent \textbf{Metrics.} We measure fidelity with PSNR, SSIM~\cite{ssim}, and LPIPS~\cite{lpips}. We evaluate video quality with VBench~\cite{vbench}, audio-video quality with T2AV-Compass~\cite{t2avcompass}, and camera pose with R.Err., T.Err., and CMC. Efficiency metrics include sparsity, FLOPs, and the latency and acceleration ratios for attention and denoising.

\par\noindent \textbf{Datasets.} For text-to-video generation, we use long text datasets expanded by GPT from the VBench team; for image-to-video generation, image-text pairs from VBench-I2V~\cite{vbenchpp}; for joint audio-video generation, text prompts from T2AV-Compass; and for world simulation, images, text, and camera trajectories from SANA-WM-Bench.

\par\noindent \textbf{Baselines.} We use FlashAttention-2~\cite{fa2} as the dense attention baseline to compare representative training-free sparse methods, including clustering-based SVG2~\cite{svg2} and SVG-EAR~\cite{svgear}, and block-based PISA~\cite{pisa} and Sol-Attn~\cite{solattn}.

\noindent FastH3 uses a single NVIDIA RTX PRO 6000 GPU, while the other models use a single NVIDIA A800 GPU. \textbf{See Appendix \ref{sec:source184} for detailed settings.}

\begin{figure*}[t!]
\centering
\includegraphics[width=\linewidth]{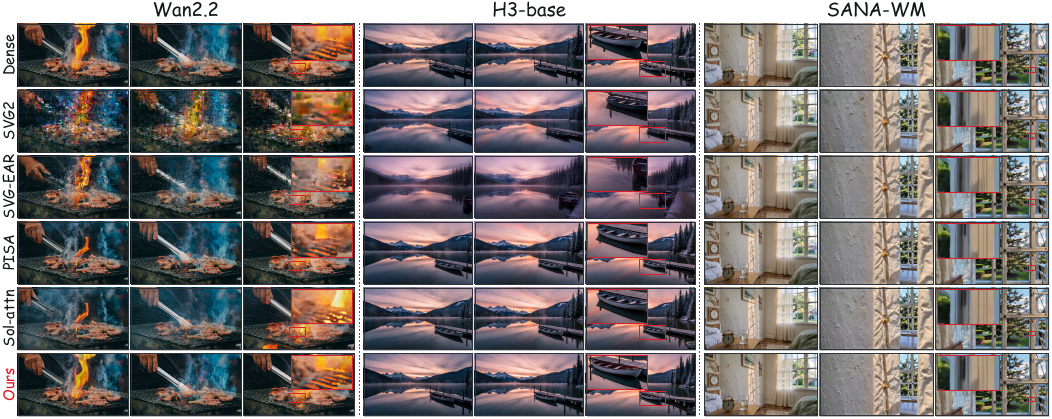}
\caption{\textbf{Visual comparison on Wan2.2, \mbox{H3-Base}, and SANA-WM.}}
\label{fig:qualitative}
\end{figure*}
\subsection{Quantitative Comparison}
\label{sec:source129}
\begin{table}[!htbp]
\centering
\caption{\textbf{Quantitative Comparison of Consistency with the Input Image, Audio and Video Quality, and Camera Pose.} Bold and underlined values indicate the best and second-best results, respectively.}
\label{tab:source133}
\begingroup
\small
\setlength{\tabcolsep}{3pt}
\begin{soltabularfit}
\begin{tabular}{l|cc|cccc|ccc}
\toprule
\multirow{3}{*}{\textbf{Method}} & \multicolumn{2}{c}{\makecell{\textbf{Consistency with}\\\textbf{the Input Image}}} & \multicolumn{4}{c}{\textbf{Audio-Visual Quality}} & \multicolumn{3}{c}{\textbf{Pose Accuracy}} \\
\cmidrule(lr){2-3}\cmidrule(lr){4-7}\cmidrule(lr){8-10}
 & \multicolumn{2}{c|}{\cellcolor{solgray}Wan2.2} & \multicolumn{4}{c|}{\cellcolor{solgray}H3-Base} & \multicolumn{3}{c}{\cellcolor{solgray}SANA-WM Refiner} \\
\cmidrule(lr){2-3}\cmidrule(lr){4-7}\cmidrule(lr){8-10}
 & \textbf{I2V-SC$\uparrow$} & \textbf{I2V-BC$\uparrow$} & \textbf{AA$\uparrow$} & \textbf{T--A$\uparrow$} & \textbf{A--V$\uparrow$} & \textbf{DS$\downarrow$} & \textbf{R.Err$\downarrow$} & \textbf{T.Err$\downarrow$} & \textbf{CMC$\downarrow$} \\
\midrule
FA & 98.72 & 99.73 & 7.150 & 0.206 & 0.252 & 0.488 & 5.847 & 0.855 & 0.893 \\
\midrule
SVG2 & 76.05 & 87.72 & \underline{6.911} & 0.213 & 0.250 & 0.815 & 6.273 & \underline{0.910} & \underline{0.949} \\
SVG-EAR & 92.74 & 96.63 & 6.783 & 0.220 & 0.260 & \underline{0.569} & 6.301 & 0.926 & 0.963 \\
PISA & 97.82 & 99.67 & 6.890 & \underline{0.240} & \textbf{0.265} & 0.677 & \underline{6.159} & 0.929 & 0.969 \\
Sol-Attn & \underline{98.43} & \underline{99.68} & 6.878 & 0.226 & 0.262 & \underline{0.569} & 6.204 & 0.911 & 0.950 \\
\rowcolor{solblue}
Sol\textsuperscript{3} & \textbf{98.53} & \textbf{99.71} & \textbf{6.986} & \textbf{0.249} & \underline{0.263} & \textbf{0.554} & \textbf{6.059} & \textbf{0.859} & \textbf{0.900} \\
\bottomrule
\end{tabular}
\end{soltabularfit}
\endgroup

\end{table}
\par\noindent \textbf{Video generation.} We evaluate video generation using text and image inputs at 480p, 540p, and 720p. Tables \ref{tab:source131} and \ref{tab:source133} report image-to-video results on Wan2.2. Compared with Sol-Attn, Sol\textsuperscript{3} increases the attention acceleration ratio from 2.45$\times$ to 3.07$\times$ while achieving the best fidelity. Appendix Table \ref{tab:source169} shows the text-to-video results using HunyuanVideo, and Appendix Tables \ref{tab:source181} and \ref{tab:source183} compare results across resolutions.

\par\noindent \textbf{Audio-Video Generation.} We further evaluate diffusion models that jointly generate RGB video and audio. Tables \ref{tab:source131} and \ref{tab:source133} show that, on H3-Base, Sol\textsuperscript{3} improves the attention acceleration ratio from 1.98$\times$ to 2.27$\times$ versus PISA and raises the audio quality metric AA from 6.890 to 6.986. Appendix Tables \ref{tab:source169} and \ref{tab:source171} report results for LTX-2.5 and for few-step generation with FastH3.

\par\noindent \textbf{World Simulation.} We use more controllable video models to extend evaluation to world simulation, where Sol\textsuperscript{3} also balances generation fidelity with camera consistency. Tables \ref{tab:source131} and \ref{tab:source133} show that, across 481 frames generated by the SANA-WM Refiner, Sol\textsuperscript{3} achieved higher generation fidelity while reducing CMC from 0.950 for Sol-Attn to 0.900 and reaching an attention acceleration ratio of up to 2.33$\times$. Comparative experiments on 961 frames appear in Appendix Tables \ref{tab:source175} and \ref{tab:source177}.

\noindent \textbf{See Appendix \ref{sec:source165} for the complete experimental results.}

\subsection{Visual Comparison}
\label{sec:source138}
Figure \ref{fig:qualitative} shows a qualitative comparison of Wan2.2, H3-Base, and SANA-WM. Most baseline models show color noise, local blurring, and detail loss, especially in texture-rich areas and fine structures, as shown in the close-up images. In contrast, Sol\textsuperscript{3}'s appearance and spatial structure more closely match full attention, preserving more complete details and contours. This reflects our core mechanism: progressively refine the scope of exact computation using criteria tailored to decision objectives at each level. \textbf{See Appendix \ref{sec:source294} for additional visual comparisons.}

\subsection{Ablation Studies}
\label{sec:source1110}
We ablate the two core components of \textbf{Sol\textsuperscript{3}}: Gaussian-guided hierarchical routing (GHR) and heterogeneous routing statistics (HRS).

\begin{table}[!htbp]
\centering
\caption{\textbf{Ablation study of hierarchical routing on FastH3.}}
\label{tab:source145}
\begingroup
\small
\setlength{\tabcolsep}{3pt}
\begin{soltabularfit}
\begin{tabular}{ccc|cccc}
\toprule
\multicolumn{3}{c|}{\textbf{GHR}} & \multirow{2}{*}{\textbf{Sparsity$\uparrow$}} & \multirow{2}{*}{\textbf{PSNR$\uparrow$}} & \multirow{2}{*}{\textbf{SSIM$\uparrow$}} & \multirow{2}{*}{\textbf{LPIPS$\downarrow$}} \\
\cmidrule(lr){1-3}
\textbf{level 1} & \textbf{level 2} & \textbf{level 3} &  &  &  &  \\
\midrule
\Checkmark & \XSolidBrush & \XSolidBrush & 83.54\% & 15.87 & 0.499 & 0.386 \\
\XSolidBrush & \Checkmark & \XSolidBrush & 85.25\% & 15.83 & 0.499 & 0.390 \\
\XSolidBrush & \XSolidBrush & \Checkmark & 85.28\% & 15.43 & 0.487 & 0.448 \\
\Checkmark & \Checkmark & \XSolidBrush & 85.22\% & 15.81 & 0.499 & 0.392 \\
\Checkmark & \XSolidBrush & \Checkmark & 84.70\% & 16.05 & 0.515 & 0.402 \\
\XSolidBrush & \Checkmark & \Checkmark & 85.09\% & 16.23 & 0.513 & 0.400 \\
\rowcolor{solblue}
\Checkmark & \Checkmark & \Checkmark & \textbf{85.37\%} & \textbf{16.75} & \textbf{0.537} & \textbf{0.366} \\
\bottomrule
\end{tabular}
\end{soltabularfit}
\endgroup

\end{table}
\par\noindent \textbf{Hierarchical routing strategy.} Table \ref{tab:source145} compares block-level routing, token-level routing, and a combination of the two. If only the first two levels are kept, the entire block must be computed exactly even when only a few tokens in a Mixed block are highly relevant to the query. Filtering directly from all candidate tokens ignores the block's overall response and gives the lowest fidelity among the variants. First, keep blocks with strong overall responses in full, then filter tokens from the remaining blocks to improve fidelity. Full routing further focuses this filtering step on Mixed blocks that are difficult to approximate, achieving the highest PSNR and SSIM and the lowest LPIPS.

\begin{table}[!htbp]
\centering
\caption{\textbf{Ablation study of heterogeneous routing statistics on \mbox{HunyuanVideo}.}}
\label{tab:source149}
\begingroup
\small
\setlength{\tabcolsep}{3pt}
\makebox[\linewidth][l]{
\begin{soltabularfit}
\begin{tabular}{l|ccc}
\toprule
\textbf{Variant} & \makecell{\textbf{Full-block}\\\textbf{selection}} & \makecell{\textbf{Approximation}\\\textbf{screening}} & \makecell{\textbf{Token}\\\textbf{comparison}} \\
\midrule
(a) & Block corr. & Block corr. & Cross-block \\
(b) & Approx. difficulty & Approx. difficulty & Cross-block \\
(c) & Block corr. & Approx. difficulty & Within-block \\
\rowcolor{solblue}
(d) & Block corr. & Approx. difficulty & Cross-block \\
\bottomrule
\end{tabular}
\end{soltabularfit}
}
\endgroup
\par\smallskip
\begingroup
\small
\setlength{\tabcolsep}{3pt}
\makebox[\linewidth][l]{
\begin{soltabularfit}
\begin{tabular}{l|ccccc}
\toprule
\textbf{Variant} & \textbf{Sparsity$\uparrow$} & \textbf{PSNR$\uparrow$} & \textbf{SSIM$\uparrow$} & \textbf{LPIPS$\downarrow$} & \textbf{Latency (s) $\downarrow$} \\
\midrule
(a) & 83.92\% & 23.76 & 0.839 & 0.115 & 1768.1 \\
(b) & 84.65\% & 23.38 & 0.823 & 0.129 & 1763.5 \\
(c) & 83.30\% & 20.42 & 0.769 & 0.181 & 1857.3 \\
\rowcolor{solblue}
(d) & \textbf{84.75\%} & \textbf{24.90} & \textbf{0.857} & \textbf{0.099} & \textbf{1750.1} \\
\bottomrule
\end{tabular}
\end{soltabularfit}
}
\endgroup

\end{table}
\par\noindent \textbf{Routing statistics.} Table \ref{tab:source149} compares scoring and filtering methods across levels under fixed hierarchical routing. When the first two levels use block-center correlation, the second level cannot distinguish blocks by representative approximation difficulty. When both use the representative approximation difficulty score, the first level no longer preserves high-response regions as a whole. Independent block-by-block screening adds higher computational overhead and may overlook members in other Mixed blocks with higher raw token relevance to the query group. In contrast, full HRS aligns scores with the decision objectives at each level and outperforms the three variant methods.

\begin{table}[!htbp]
\centering
\caption{\textbf{Generated fidelity and filtering overhead of FastH3 using different threshold rules at three sparsity levels.}}
\label{tab:source154}
\begingroup
\small
\setlength{\tabcolsep}{3pt}
\begin{soltabularfit}
\begin{tabular}{l|cccccc}
\toprule
\textbf{Method} & \textbf{Sparsity$\uparrow$} & \textbf{PSNR$\uparrow$} & \textbf{SSIM$\uparrow$} & \textbf{LPIPS$\downarrow$} & \textbf{Selection Time(ms)$\downarrow$} & \textbf{Overhead $\downarrow$} \\
\midrule
Top-K & 75.33\% & 17.44 & 0.564 & 0.314 & 735.9 & 172.9\% \\
Top-p & 75.31\% & 16.08 & 0.500 & 0.386 & 957.0 & 224.9\% \\
\rowcolor{solblue}
Gaussian & \textbf{75.63\%} & \textbf{17.72} & \textbf{0.572} & \textbf{0.310} & \textbf{30.73} & \textbf{7.220\%} \\
\midrule
Top-K & 80.35\% & 16.96 & 0.548 & 0.335 & 732.9 & 172.2\% \\
Top-p & 80.43\% & 15.68 & 0.486 & 0.404 & 954.8 & 224.3\% \\
\rowcolor{solblue}
Gaussian & \textbf{80.58\%} & \textbf{17.28} & \textbf{0.556} & \textbf{0.333} & \textbf{30.72} & \textbf{7.220\%} \\
\midrule
Top-K & 85.28\% & 16.55 & 0.529 & 0.372 & 731.7 & 171.9\% \\
Top-p & 85.25\% & 14.99 & 0.458 & 0.438 & 956.0 & 224.6\% \\
\rowcolor{solblue}
Gaussian & \textbf{85.37\%} & \textbf{16.75} & \textbf{0.537} & \textbf{0.366} & \textbf{30.78} & \textbf{7.230\%} \\
\bottomrule
\end{tabular}
\end{soltabularfit}
\endgroup

\end{table}
\par\noindent \textbf{Screening rules.} Table \ref{tab:source154} compares three screening rules on FastH3. Top-K keeps a fixed share of high-score items for each query group. Top-p stops screening once cumulative weight reaches a threshold; in groups with concentrated weights, only a few candidates remain, so some interactions that should be computed exactly are approximated. Gaussian screening sets a threshold using the mean and standard deviation of scores within a group, so different query groups retain different shares of candidates. Each score is compared only with the group's threshold, so sorting is unnecessary. Across three sparsity levels, Gaussian filtering achieved the best generation fidelity and the highest selection-phase acceleration ratios compared with Top-K and Top-p: 23.95$\times$ and 31.14$\times$, respectively.

\begin{table}[!htbp]
\centering
\caption{\textbf{Sensitivity of the three-level threshold parameters on \mbox{H3-Base}.}}
\label{tab:source158}
\begingroup
\small
\setlength{\tabcolsep}{3pt}
\newcommand{\solparamtriplet}[3]{$(\phantom{1.775}\llap{$#1$},\phantom{-0.125}\llap{$#2$},\phantom{0.5625}\llap{$#3$})$}
\begin{soltabularfit}
\begin{tabular}{l|cccc}
\toprule
{\bfseries\boldmath\solparamtriplet{\beta_1}{\beta_2}{\beta_3}} & \textbf{Sparsity$\uparrow$} & \textbf{PSNR$\uparrow$} & \textbf{SSIM$\uparrow$} & \textbf{LPIPS$\downarrow$} \\
\midrule
\multicolumn{1}{c|}{Sol-Attn~\cite{solattn}} & 78.36\% & 16.06 & 0.570 & 0.384 \\
\midrule
\rowcolor{solblue}
\solparamtriplet{1.9}{0}{0.625} & 80.78\% & \textbf{16.82} & \textbf{0.598} & 0.348 \\
\solparamtriplet{1.775}{0}{0.625} & 79.75\% & 16.77 & 0.593 & \textbf{0.344} \\
\solparamtriplet{2.025}{0}{0.625} & 81.67\% & 16.19 & 0.587 & 0.377 \\
\solparamtriplet{1.9}{-0.125}{0.625} & 79.47\% & 16.27 & 0.591 & 0.349 \\
\solparamtriplet{1.9}{0.125}{0.625} & \textbf{82.12\%} & 16.38 & 0.592 & 0.375 \\
\solparamtriplet{1.9}{0}{0.5625} & 79.85\% & 16.78 & 0.587 & 0.351 \\
\solparamtriplet{1.9}{0}{0.6875} & 81.63\% & 16.31 & 0.590 & 0.360 \\
\bottomrule
\end{tabular}
\end{soltabularfit}
\endgroup

\end{table}
\par\noindent \textbf{Threshold parameters.} Table \ref{tab:source158} tests the sensitivity of the three-level routing thresholds in H3-Base. After adjusting any level up or down around the default values, all six configurations still outperformed the baseline Sol-Attn on fidelity metrics: PSNR ranged from 16.19 to 16.78 dB, SSIM from 0.587 to 0.593, and LPIPS from 0.344 to 0.377. This stable range shows that hierarchical routing's fidelity advantage does not depend on a precisely tuned threshold set. The default configuration (1.9, 0, 0.625) gives the most balanced performance across all three fidelity metrics.

\section{Conclusion}
\label{sec:source161}
This paper proposes Sol\textsuperscript{3}, a training-free hierarchical sparse attention method for accelerating diffusion video models. To address Sol-Attn's block-level decision-making limits, Sol\textsuperscript{3} uses Gaussian-guided hierarchical routing to refine computation from blocks to tokens on demand, and uses heterogeneous routing statistics to guide approximation and member selection during refinement. Experiments in video generation, joint audio-video generation, and world simulation tasks demonstrate that Sol\textsuperscript{3} achieves up to 86\% attention sparsity and 3.07$\times$ attention acceleration while maintaining generation quality. We hope that Sol\textsuperscript{3} will provide an effective, practical approach to achieving efficient attention mechanisms in video diffusion models.

\clearpage
\section*{LLM Usage Statement}
Throughout the research and paper-writing process, we used OpenAI Codex to help edit text, modify images, organize and check mathematical derivations in the appendices, and generate, modify, and debug code. We reviewed all AI-assisted content and assume full responsibility for the final content, code, and deliverables.

{
\raggedbottom
\small
\bibliographystyle{ieeenat_fullname}
\bibliography{main}
\clearpage
}
\clearpage
\maketitlesupplementary
\appendix
\numberwithin{figure}{section}
\numberwithin{table}{section}
\numberwithin{equation}{section}
\renewcommand{\thefigure}{\thesection\arabic{figure}}
\renewcommand{\thetable}{\thesection\arabic{table}}
\section{Complete Experimental Results}
\label{sec:source165}
\begin{table*}[t!]
\centering
\caption{\textbf{Quantitative Comparison Across Different Diffusion Models.} Values outside and inside parentheses report attention and denoising speedups, respectively. Bold and underlined values indicate the best and second-best results, respectively.}
\label{tab:source169}
\begingroup
\small
\setlength{\tabcolsep}{3pt}
\begin{soltabularfit}
\begin{tabular}{l|ccccccccc}
\toprule
\multirow{2}{*}{\textbf{Method}} & \multicolumn{3}{c}{\textbf{Similarity}} & \multicolumn{3}{c}{\textbf{VBench}} & \multicolumn{3}{c}{\textbf{Efficiency}} \\
\cmidrule(lr){2-4}\cmidrule(lr){5-7}\cmidrule(lr){8-10}
 & \textbf{PSNR$\uparrow$} & \textbf{SSIM$\uparrow$} & \textbf{LPIPS$\downarrow$} & \textbf{ImgQual$\uparrow$} & \textbf{SubCons$\uparrow$} & \textbf{AVG$\uparrow$} & \textbf{Sparsity$\uparrow$} & \textbf{FLOPs(P)$\downarrow$} & \textbf{Speedup$\uparrow$} \\
\midrule
\multicolumn{10}{c}{\cellcolor{solgray}HunyuanVideo (13B, Text-to-Video, 720p, 129 Frames)} \\
\midrule
FA & $\infty$ & 1.000 & 0.000 & 71.30 & 97.27 & 73.66 & 0.000\% & 602.32 & 1.00$\times$ (1.00$\times$) \\
\midrule
SVG2 & 22.16 & 0.741 & 0.293 & 58.73 & 95.56 & 65.60 & 71.07\% & 302.06 & 2.50$\times$ (1.63$\times$) \\
SVG-EAR & 22.81 & 0.770 & 0.249 & 62.77 & \underline{97.53} & 67.41 & \underline{83.90\%} & \underline{247.83} & \underline{2.77$\times$} (\underline{1.69$\times$}) \\
PISA & \underline{23.87} & \underline{0.836} & \underline{0.124} & \underline{73.08} & 97.51 & \underline{72.93} & 83.56\% & 249.19 & 2.41$\times$ (1.62$\times$) \\
Sol-Attn & 22.60 & 0.796 & 0.139 & \textbf{73.24} & 97.44 & 72.89 & 82.26\% & 254.76 & 2.25$\times$ (1.57$\times$) \\
\rowcolor{solblue}
Sol\textsuperscript{3} & \textbf{24.90} & \textbf{0.857} & \textbf{0.099} & 72.50 & \textbf{97.78} & \textbf{73.83} & \textbf{84.75\%} & \textbf{244.24} & {\bfseries\boldmath 2.84$\times$} ({\bfseries\boldmath 1.85$\times$}) \\
\midrule
\multicolumn{10}{c}{\cellcolor{solgray}LTX-2.5 (22B, Text-to-Audio-Video, 1024p, 361 Frames)} \\
\midrule
FA & $\infty$ & 1.000 & 0.000 & 64.45 & 86.23 & 77.19 & 0.000\% & 86.339 & 1.00$\times$ (1.00$\times$) \\
\midrule
SVG2 & 15.68 & 0.686 & 0.516 & 56.34 & 83.72 & \underline{76.78} & 77.18\% & 66.417 & 1.63$\times$ (1.11$\times$) \\
SVG-EAR & 17.37 & 0.711 & 0.431 & 59.32 & \textbf{86.11} & 76.62 & 70.77\% & 68.074 & 1.25$\times$ (1.07$\times$) \\
PISA & 17.32 & 0.715 & 0.419 & \textbf{64.34} & 85.17 & 76.51 & \underline{80.07\%} & \underline{65.675} & 1.76$\times$ (1.12$\times$) \\
Sol-Attn & \underline{17.44} & \underline{0.718} & \underline{0.414} & 62.63 & \underline{85.38} & 76.12 & 79.81\% & 65.740 & \underline{2.34$\times$} (\underline{1.20$\times$}) \\
\rowcolor{solblue}
Sol\textsuperscript{3} & \textbf{17.87} & \textbf{0.735} & \textbf{0.395} & \underline{63.15} & 85.19 & \textbf{76.93} & \textbf{85.68\%} & \textbf{64.225} & {\bfseries\boldmath 2.58$\times$} ({\bfseries\boldmath 1.22$\times$}) \\
\midrule
\multicolumn{10}{c}{\cellcolor{solgray}FastH3 (33B, Text-to-Audio-Video, 768p, 243 Frames)} \\
\midrule
FA & $\infty$ & 1.000 & 0.000 & 72.31 & 84.94 & 77.54 & 0.000\% & 42.393 & 1.00$\times$ (1.00$\times$) \\
\midrule
SVG2 & 15.31 & 0.453 & 0.568 & 61.83 & 83.92 & 74.66 & 78.03\% & 25.071 & 2.62$\times$ (\underline{1.48$\times$}) \\
SVG-EAR & 15.94 & 0.487 & 0.453 & 67.91 & 84.26 & 76.84 & 74.40\% & 25.804 & 2.31$\times$ (1.44$\times$) \\
PISA & 16.15 & 0.513 & 0.384 & 71.49 & \textbf{85.51} & 77.31 & \underline{81.11\%} & \underline{24.321} & 2.03$\times$ (1.39$\times$) \\
Sol-Attn & \underline{16.25} & \underline{0.527} & \underline{0.373} & \underline{71.50} & \underline{85.24} & \underline{77.37} & 74.79\% & 25.727 & \underline{2.64$\times$} (\underline{1.48$\times$}) \\
\rowcolor{solblue}
Sol\textsuperscript{3} & \textbf{16.75} & \textbf{0.537} & \textbf{0.366} & \textbf{72.05} & 85.21 & \textbf{77.48} & \textbf{85.37\%} & \textbf{23.360} & {\bfseries\boldmath 2.72$\times$} ({\bfseries\boldmath 1.50$\times$}) \\
\bottomrule
\end{tabular}
\end{soltabularfit}
\endgroup

\end{table*}
\subsection{Additional Diffusion Models}
\label{sec:source166}
\begin{table}[!htbp]
\centering
\caption{\textbf{Quality Comparison Across Different Audio-Video Models.} Bold and underlined values indicate the best and second-best results, respectively.}
\label{tab:source171}
\begingroup
\small
\setlength{\tabcolsep}{3pt}
\begin{soltabularfit}
\begin{tabular}{l|cccc|cccc}
\toprule
\multirow{3}{*}{\textbf{Method}} & \multicolumn{4}{c}{\textbf{Audio-Visual Quality}} & \multicolumn{4}{c}{\textbf{Audio-Visual Quality}} \\
\cmidrule(lr){2-5}\cmidrule(lr){6-9}
 & \multicolumn{4}{c|}{\cellcolor{solgray}LTX-2.5 (1024p, 361 Frames)} & \multicolumn{4}{c}{\cellcolor{solgray}FastH3 (768p, 243 Frames)} \\
\cmidrule(lr){2-5}\cmidrule(lr){6-9}
 & \textbf{AA$\uparrow$} & \textbf{T--A$\uparrow$} & \textbf{A--V$\uparrow$} & \textbf{DS$\downarrow$} & \textbf{AA$\uparrow$} & \textbf{T--A$\uparrow$} & \textbf{A--V$\uparrow$} & \textbf{DS$\downarrow$} \\
\midrule
FA & 6.405 & 0.270 & 0.190 & 0.971 & 6.995 & 0.237 & 0.219 & 0.703 \\
\midrule
SVG2 & 6.308 & \textbf{0.282} & 0.192 & 0.913 & 6.776 & 0.235 & 0.211 & 0.873 \\
SVG-EAR & 6.311 & 0.243 & 0.170 & 0.850 & 6.936 & \underline{0.240} & 0.214 & 0.780 \\
PISA & 6.396 & 0.258 & \underline{0.217} & \textbf{0.700} & \underline{6.956} & 0.239 & \underline{0.220} & \underline{0.618} \\
Sol-Attn & \underline{6.446} & 0.266 & 0.214 & 0.850 & 6.948 & 0.238 & 0.219 & 0.671 \\
\rowcolor{solblue}
Sol\textsuperscript{3} & \textbf{6.592} & \underline{0.273} & \textbf{0.219} & \underline{0.775} & \textbf{6.962} & \textbf{0.242} & \textbf{0.222} & \textbf{0.614} \\
\bottomrule
\end{tabular}
\end{soltabularfit}
\endgroup

\end{table}
We provide the complete comparison results for the HunyuanVideo, LTX-2.5, and FastH3 experiments described in the main text in Tables \ref{tab:source169} and \ref{tab:source171}.

\subsection{Longer Videos}
\label{sec:source172}
We provide the comparison results for longer generated sequences in Tables \ref{tab:source175} and \ref{tab:source177}.

\begin{table*}[t!]
\centering
\caption{\textbf{Quantitative Comparison for Longer Frame Counts.} Values outside and inside parentheses report attention and denoising speedups, respectively. Bold and underlined values indicate the best and second-best results, respectively.}
\label{tab:source175}
\begingroup
\small
\setlength{\tabcolsep}{3pt}
\begin{soltabularfit}
\begin{tabular}{l|ccccccccc}
\toprule
\multirow{2}{*}{\textbf{Method}} & \multicolumn{3}{c}{\textbf{Similarity}} & \multicolumn{3}{c}{\textbf{VBench}} & \multicolumn{3}{c}{\textbf{Efficiency}} \\
\cmidrule(lr){2-4}\cmidrule(lr){5-7}\cmidrule(lr){8-10}
 & \textbf{PSNR$\uparrow$} & \textbf{SSIM$\uparrow$} & \textbf{LPIPS$\downarrow$} & \textbf{ImgQual$\uparrow$} & \textbf{SubCons$\uparrow$} & \textbf{AVG$\uparrow$} & \textbf{Sparsity$\uparrow$} & \textbf{FLOPs(P)$\downarrow$} & \textbf{Speedup$\uparrow$} \\
\midrule
\multicolumn{10}{c}{\cellcolor{solgray}H3-Base (33B, Text-to-Audio-Video, 768p, 243 Frames)} \\
\midrule
FA & $\infty$ & 1.000 & 0.000 & 66.91 & 88.56 & 77.12 & 0.000\% & 518.76 & 1.00$\times$ (1.00$\times$) \\
\midrule
SVG2 & \underline{15.86} & 0.538 & 0.423 & 59.11 & 87.57 & 76.31 & 30.05\% & 420.93 & 1.03$\times$ (0.90$\times$) \\
SVG-EAR & 15.18 & 0.510 & 0.501 & 52.66 & 85.19 & 72.13 & 27.94\% & 427.80 & 1.11$\times$ (0.99$\times$) \\
PISA & 15.81 & \underline{0.544} & \underline{0.377} & \underline{67.29} & \underline{88.70} & \underline{77.26} & \underline{81.07\%} & \underline{254.79} & 2.09$\times$ (1.47$\times$) \\
Sol-Attn & 15.44 & 0.534 & 0.379 & \textbf{67.60} & 87.57 & 77.22 & 78.93\% & 262.36 & \underline{2.15$\times$} (\underline{1.48$\times$}) \\
\rowcolor{solblue}
Sol\textsuperscript{3} & \textbf{16.21} & \textbf{0.575} & \textbf{0.348} & 66.43 & \textbf{88.85} & \textbf{77.36} & \textbf{81.36\%} & \textbf{254.45} & {\bfseries\boldmath 2.45$\times$} ({\bfseries\boldmath 1.65$\times$}) \\
\midrule
\multicolumn{10}{c}{\cellcolor{solgray}SANA-WM Refiner (17B, Video-to-Video, 720p, 961 Frames)} \\
\midrule
FA & $\infty$ & 1.000 & 0.000 & 73.21 & 81.31 & 76.67 & 0.000\% & 33.820 & 1.00$\times$ (1.00$\times$) \\
\midrule
SVG2 & 20.81 & 0.713 & 0.183 & 72.94 & 80.94 & 76.49 & 69.93\% & 21.709 & \underline{2.41$\times$} (\underline{1.41$\times$}) \\
SVG-EAR & 22.31 & 0.771 & 0.139 & 72.45 & 81.01 & 76.46 & 69.85\% & 21.723 & 2.26$\times$ (1.38$\times$) \\
PISA & \underline{23.77} & \underline{0.823} & \underline{0.106} & 72.75 & \textbf{81.47} & 76.61 & \underline{79.03\%} & \underline{20.139} & 2.10$\times$ (1.35$\times$) \\
Sol-Attn & 23.63 & 0.818 & 0.107 & \underline{73.13} & 81.18 & \underline{76.82} & 77.90\% & 20.329 & 2.22$\times$ (1.37$\times$) \\
\rowcolor{solblue}
Sol\textsuperscript{3} & \textbf{23.95} & \textbf{0.827} & \textbf{0.101} & \textbf{73.26} & \underline{81.39} & \textbf{76.91} & \textbf{79.14\%} & \textbf{20.114} & {\bfseries\boldmath 2.46$\times$} ({\bfseries\boldmath 1.46$\times$}) \\
\midrule
\multicolumn{10}{c}{\cellcolor{solgray}LTX-2.5 (22B, Text-to-Audio-Video, 1024p, 481 Frames)} \\
\midrule
FA & $\infty$ & 1.000 & 0.000 & 64.23 & 85.87 & 75.39 & 0.000\% & 127.21 & 1.00$\times$ (1.00$\times$) \\
\midrule
SVG2 & 15.94 & 0.644 & 0.542 & 57.36 & 82.93 & 74.67 & 77.01\% & 92.255 & 1.99$\times$ (1.18$\times$) \\
SVG-EAR & \underline{17.77} & 0.698 & 0.413 & 61.79 & \textbf{85.60} & 76.23 & 70.62\% & 95.157 & 1.53$\times$ (1.12$\times$) \\
PISA & 17.73 & 0.680 & 0.419 & \underline{64.16} & 83.72 & 76.18 & \underline{80.05\%} & \underline{90.880} & 1.82$\times$ (1.16$\times$) \\
Sol-Attn & 17.75 & \underline{0.700} & \underline{0.409} & 63.93 & 83.73 & \underline{76.37} & 79.86\% & 90.964 & \underline{2.21$\times$} (\underline{1.19$\times$}) \\
\rowcolor{solblue}
Sol\textsuperscript{3} & \textbf{17.82} & \textbf{0.714} & \textbf{0.405} & \textbf{64.27} & \underline{84.73} & \textbf{76.76} & \textbf{85.88\%} & \textbf{88.230} & {\bfseries\boldmath 2.83$\times$} ({\bfseries\boldmath 1.28$\times$}) \\
\midrule
\multicolumn{10}{c}{\cellcolor{solgray}FastH3 (33B, Text-to-Audio-Video, 768p, 345 Frames)} \\
\midrule
FA & $\infty$ & 1.000 & 0.000 & 72.49 & 86.32 & 76.50 & 0.000\% & 78.200 & 1.00$\times$ (1.00$\times$) \\
\midrule
SVG2 & 15.97 & 0.496 & 0.551 & 61.65 & 85.28 & 71.91 & 77.78\% & 43.547 & 2.69$\times$ (\underline{1.58$\times$}) \\
SVG-EAR & 16.36 & 0.518 & 0.448 & 68.34 & 85.99 & 76.12 & 74.38\% & 45.061 & 2.50$\times$ (1.54$\times$) \\
PISA & 16.62 & 0.524 & 0.360 & \textbf{72.48} & 86.06 & 76.33 & \underline{81.00\%} & \underline{42.064} & 2.09$\times$ (1.44$\times$) \\
Sol-Attn & \underline{16.83} & \underline{0.544} & \underline{0.329} & \underline{72.03} & \underline{86.15} & \underline{76.39} & 74.77\% & 44.863 & \underline{2.71$\times$} (\underline{1.58$\times$}) \\
\rowcolor{solblue}
Sol\textsuperscript{3} & \textbf{17.07} & \textbf{0.549} & \textbf{0.308} & 71.90 & \textbf{86.52} & \textbf{76.48} & \textbf{85.53\%} & \textbf{40.071} & {\bfseries\boldmath 2.80$\times$} ({\bfseries\boldmath 1.61$\times$}) \\
\bottomrule
\end{tabular}
\end{soltabularfit}
\endgroup

\end{table*}
\begin{table*}[t!]
\centering
\caption{\textbf{Task metrics for longer generated sequences.} Bold and underlined values indicate the best and second-best results, respectively.}
\label{tab:source177}
\begingroup
\small
\setlength{\tabcolsep}{3pt}
\begin{soltabularfit}
\begin{tabular}{l|cccc|cccc|cccc|ccc}
\toprule
\multirow{3}{*}{\textbf{Method}} & \multicolumn{12}{c}{\textbf{Audio-Visual Quality}} & \multicolumn{3}{c}{\textbf{Pose Accuracy}} \\
\cmidrule(lr){2-13}\cmidrule(lr){14-16}
 & \multicolumn{4}{c|}{\cellcolor{solgray}H3-Base (768p, 243 Frames)} & \multicolumn{4}{c|}{\cellcolor{solgray}LTX-2.5 (1024p, 481 Frames)} & \multicolumn{4}{c|}{\cellcolor{solgray}FastH3 (768p, 345 Frames)} & \multicolumn{3}{c}{\cellcolor{solgray}SANA-WM Refiner (720p, 961 Frames)} \\
\cmidrule(lr){2-5}\cmidrule(lr){6-9}\cmidrule(lr){10-13}\cmidrule(lr){14-16}
 & \textbf{AA$\uparrow$} & \textbf{T--A$\uparrow$} & \textbf{A--V$\uparrow$} & \textbf{DS$\downarrow$} & \textbf{AA$\uparrow$} & \textbf{T--A$\uparrow$} & \textbf{A--V$\uparrow$} & \textbf{DS$\downarrow$} & \textbf{AA$\uparrow$} & \textbf{T--A$\uparrow$} & \textbf{A--V$\uparrow$} & \textbf{DS$\downarrow$} & \textbf{R.Err$\downarrow$} & \textbf{T.Err$\downarrow$} & \textbf{CMC$\downarrow$} \\
\midrule
FA & 7.205 & 0.249 & 0.211 & 1.050 & 6.347 & 0.240 & 0.203 & 0.974 & 6.639 & 0.261 & 0.217 & 0.778 & 11.84 & 1.692 & 1.767 \\
\midrule
SVG2 & 6.983 & 0.226 & 0.216 & 0.771 & 6.249 & 0.199 & 0.162 & 1.120 & 6.520 & \textbf{0.278} & 0.212 & 0.888 & 14.28 & 1.851 & 1.950 \\
SVG-EAR & 7.095 & 0.238 & \underline{0.226} & 0.886 & \textbf{6.502} & \underline{0.224} & 0.183 & \underline{1.093} & 6.537 & \underline{0.276} & 0.213 & \underline{0.810} & 12.97 & 1.743 & 1.825 \\
PISA & 6.957 & 0.224 & 0.222 & 0.971 & 6.431 & 0.209 & 0.181 & 1.267 & \textbf{6.635} & 0.273 & 0.212 & 0.844 & 12.66 & \underline{1.724} & \underline{1.806} \\
Sol-Attn & \underline{7.274} & \textbf{0.260} & 0.203 & \underline{0.714} & 6.407 & 0.188 & \underline{0.186} & \textbf{1.000} & 6.601 & 0.269 & \underline{0.214} & 0.849 & \underline{10.95} & 1.795 & 1.853 \\
\rowcolor{solblue}
Sol\textsuperscript{3} & \textbf{7.313} & \underline{0.243} & \textbf{0.231} & \textbf{0.686} & \underline{6.462} & \textbf{0.225} & \textbf{0.198} & 1.213 & \underline{6.606} & \textbf{0.278} & \textbf{0.218} & \textbf{0.807} & \textbf{10.85} & \textbf{1.718} & \textbf{1.801} \\
\bottomrule
\end{tabular}
\end{soltabularfit}
\endgroup

\end{table*}
\begin{table*}[t!]
\centering
\caption{\textbf{Quantitative comparison across different resolutions.} Values outside and inside parentheses report attention and denoising speedups, respectively. Bold and underlined values indicate the best and second-best results, respectively.}
\label{tab:source181}
\begingroup
\small
\setlength{\tabcolsep}{3pt}
\begin{soltabularfit}
\begin{tabular}{l|ccccccccc}
\toprule
\multirow{2}{*}{\textbf{Method}} & \multicolumn{3}{c}{\textbf{Similarity}} & \multicolumn{3}{c}{\textbf{VBench}} & \multicolumn{3}{c}{\textbf{Efficiency}} \\
\cmidrule(lr){2-4}\cmidrule(lr){5-7}\cmidrule(lr){8-10}
 & \textbf{PSNR$\uparrow$} & \textbf{SSIM$\uparrow$} & \textbf{LPIPS$\downarrow$} & \textbf{ImgQual$\uparrow$} & \textbf{SubCons$\uparrow$} & \textbf{AVG$\uparrow$} & \textbf{Sparsity$\uparrow$} & \textbf{FLOPs(P)$\downarrow$} & \textbf{Speedup$\uparrow$} \\
\midrule
\multicolumn{10}{c}{\cellcolor{solgray}Wan2.2 (14B, Image-to-Video, 480p, 81 Frames)} \\
\midrule
FA & $\infty$ & 1.000 & 0.000 & 69.83 & 96.44 & 71.31 & 0.000\% & 134.27 & 1.00$\times$ (1.00$\times$) \\
\midrule
SVG2 & 17.20 & 0.420 & 0.564 & 51.03 & 70.73 & 56.24 & 80.64\% & 87.263 & 2.20$\times$ (1.25$\times$) \\
SVG-EAR & 22.05 & 0.675 & 0.208 & 70.60 & 93.52 & 68.67 & 84.48\% & 85.027 & 2.49$\times$ (1.29$\times$) \\
PISA & 24.68 & 0.780 & 0.080 & 70.28 & 95.51 & 71.66 & \underline{85.16\%} & \underline{84.630} & \underline{2.58$\times$} (\underline{1.31$\times$}) \\
Sol-Attn & \underline{24.76} & \underline{0.799} & \underline{0.071} & \textbf{71.10} & \underline{95.80} & \underline{72.13} & 84.52\% & 85.001 & 2.14$\times$ (1.26$\times$) \\
\rowcolor{solblue}
Sol\textsuperscript{3} & \textbf{24.98} & \textbf{0.812} & \textbf{0.060} & \underline{71.01} & \textbf{95.91} & \textbf{72.84} & \textbf{86.69\%} & \textbf{83.735} & {\bfseries\boldmath 2.73$\times$} ({\bfseries\boldmath 1.53$\times$}) \\
\midrule
\multicolumn{10}{c}{\cellcolor{solgray}HunyuanVideo (13B, Text-to-Video, 540p, 129 Frames)} \\
\midrule
FA & $\infty$ & 1.000 & 0.000 & 69.82 & 97.72 & 73.57 & 0.000\% & 213.65 & 1.00$\times$ (1.00$\times$) \\
\midrule
SVG2 & 22.44 & 0.756 & 0.227 & 63.07 & 96.60 & 68.61 & 73.70\% & 113.67 & 2.47$\times$ (1.52$\times$) \\
SVG-EAR & \underline{23.25} & \underline{0.812} & 0.169 & 63.88 & 96.82 & 71.13 & 83.43\% & 100.36 & \underline{2.73$\times$} (\underline{1.59$\times$}) \\
PISA & 22.90 & 0.797 & \underline{0.139} & \underline{69.46} & \textbf{97.52} & \underline{72.84} & \underline{83.57\%} & \underline{100.24} & 2.42$\times$ (1.51$\times$) \\
Sol-Attn & 22.28 & 0.780 & 0.144 & 69.29 & \underline{97.34} & 72.65 & 83.10\% & 100.90 & 2.24$\times$ (1.47$\times$) \\
\rowcolor{solblue}
Sol\textsuperscript{3} & \textbf{23.94} & \textbf{0.827} & \textbf{0.124} & \textbf{69.52} & 96.92 & \textbf{72.98} & \textbf{83.79\%} & \textbf{99.935} & {\bfseries\boldmath 2.77$\times$} ({\bfseries\boldmath 1.73$\times$}) \\
\bottomrule
\end{tabular}
\end{soltabularfit}
\endgroup

\end{table*}
\subsection{Different Resolutions}
\label{sec:source178}
\begin{table}[!htbp]
\centering
\caption{\textbf{Consistency with the Input Image at Different Resolutions.} Bold and underlined values indicate the best and second-best results, respectively.}
\label{tab:source183}
\begingroup
\small
\setlength{\tabcolsep}{3pt}
\begin{soltabularfit}
\begin{tabular}{l|cc}
\toprule
\multirow{3}{*}{\textbf{Method}} & \multicolumn{2}{c}{\textbf{Consistency with the Input Image}} \\
\cmidrule(lr){2-3}
 & \multicolumn{2}{c}{\cellcolor{solgray}Wan2.2 (480p, 81 Frames)} \\
\cmidrule(lr){2-3}
 & \textbf{I2V-SC$\uparrow$} & \textbf{I2V-BC$\uparrow$} \\
\midrule
FA & 98.39 & 99.29 \\
\midrule
SVG2 & 85.29 & 85.30 \\
SVG-EAR & 93.60 & 96.57 \\
PISA & 98.04 & 99.01 \\
Sol-Attn & \underline{98.22} & \underline{99.03} \\
\rowcolor{solblue}
Sol\textsuperscript{3} & \textbf{98.41} & \textbf{99.09} \\
\bottomrule
\end{tabular}
\end{soltabularfit}
\endgroup

\end{table}
In Tables \ref{tab:source181} and \ref{tab:source183}, we provide the complete comparison results for Wan2.2 and HunyuanVideo at different resolutions, as discussed in the main text.

\section{Experimental Details}
\label{sec:source184}
\subsection{Model and Generation Settings}
\label{sec:source185}
We evaluated HunyuanVideo, Wan2.2, H3-Base, FastH3, LTX-2.5, and SANA-WM on text-to-video, image-to-video, audio-video co-generation, and camera-conditional world modeling. FastH3 used one NVIDIA RTX PRO 6000 GPU; the other models used one NVIDIA A800 GPU.

In the video generation experiments, HunyuanVideo generated 129 frames, and Wan2.2 generated 81 frames. The main text reports Wan2.2 at 720p; Appendix \ref{sec:source166} adds HunyuanVideo at 720p, and Appendix \ref{sec:source178} adds HunyuanVideo at 540p and Wan2.2 at 480p.

In the text-to-audio and video generation experiments, H3-Base and FastH3 used 768p, while LTX-2.5 used 1024p. The main text reports H3-Base generating 124 frames; Appendix \ref{sec:source166} adds FastH3 generating 243 frames and LTX-2.5 generating 361 frames. Appendix \ref{sec:source172} further reports H3-Base at 243 frames, FastH3 at 345 frames, and LTX-2.5 at 481 frames.

In the world modeling experiment, SANA-WM Refiner used 720p. The main text reports 481 frames, and Appendix \ref{sec:source172} adds 961 frames.

\subsection{Sparse Attention Settings}
\label{sec:source190}
All sparse methods on the same model use the same time-based warm-up and layer-protection strategies. Wan2.2 and HunyuanVideo perform 40 and 50 denoising steps, respectively, with full attention in the first 6 and 8 steps. H3-Base performs 49 denoising steps, with full attention in the first 5 steps; FastH3 performs 4 denoising steps, with full attention in the first step.

LTX-2.5's sparse replacement is applied to Stage-1, which performs 30 denoising steps, with full attention in the first 3 steps; Stage-2 keeps the original computations. SANA-WM Refiner performs 3 denoising steps. Full attention is used in the first step, while sparse computations are enabled for video self-attention in the remaining two steps.

All models retain full attention in the first layer; HunyuanVideo retains the first layer of both the dual-stream and single-stream modules.

\subsection{Evaluation Metrics}
\label{sec:source193}
\subsubsection{Generation Quality}
\label{sec:source194}
\par\noindent \textbf{Fidelity.} PSNR, SSIM, and LPIPS measure how closely the generated results match the full attention output.

\par\noindent \textbf{Video Quality.} We use VBench to evaluate video quality. ImgQual and SubCons indicate imaging quality and subject consistency, respectively; AVG is the arithmetic mean of eight metrics: subject consistency, background consistency, temporal flickering, motion smoothness, aesthetic quality, imaging quality, dynamic degree, and overall consistency. The Video-Image Subject Consistency (I2V-SC) and Video-Image Background Consistency (I2V-BC) metrics for the image-to-video generation task are computed on the corresponding VBench-I2V subsets.

\par\noindent \textbf{Audio-Video Quality.} For audio-video tasks, we report audio quality (AA), text--audio consistency (T--A), audio--video consistency (A--V), and synchronization offset (DS). AA is the average of AudioBox~\cite{audiobox}'s PQ and CU scores; T--A uses CLAP~\cite{clap}; A--V uses ImageBind~\cite{imagebind}; and DS is the absolute value of the synchronization offset predicted by Synchformer~\cite{synchformer}.

\par\noindent \textbf{World Model Quality.} SANA-WM uses the pose evaluation workflow provided by Pi3X~\cite{pi3} and the benchmark, reporting rotation error (R.Err.), translation error (T.Err.), and camera motion consistency (CMC).

\subsubsection{Efficiency Metrics}
\label{sec:source199}
\par\noindent \textbf{Sparsity rate.} The sparsity metric measures the proportion of video query--key interactions for which exact computation is skipped in calls that use sparse routing:

\begin{equation}
S_{\mathrm{VV}}
=1-\frac{\sum_{c\in\mathcal C_{\mathrm{sparse}}}N^{\mathrm{exact}}_{\mathrm{VV},c}}
{\sum_{c\in\mathcal C_{\mathrm{sparse}}}N^{\mathrm{all}}_{\mathrm{VV},c}},
\label{eq:source201}
\end{equation}
The numerator represents the total number of video interaction pairs computed exactly, and the denominator represents the total number of all possible video interaction pairs in those calls.

\par\noindent \textbf{Attention Speedup.} Attn Speedup is the ratio of total attention time for a paired FA to the corresponding method, measured across sparse computation steps and layers. The timing range runs from after QKV projection and position encoding to before output projection, and includes routing, reordering, indexing, exact computation, and representative term compensation and merging. In Appendix \ref{sec:source1503}, the fused input preparation operator runs before timing begins, and the fused residual update operator runs after timing ends.

\par\noindent \textbf{Denoising Efficiency.} Latency is the denoising time for the stage under test, while Speedup is the ratio of the paired FA to the total denoising time of the corresponding method at that stage. Both fusion operators' processing times are included in Sol\textsuperscript{3}'s denoising time. For LTX-2.5 and SANA-WM, the denoising times for Stage-1 and the Refiner, respectively, are calculated.

\par\noindent \textbf{Computational load.} FLOPs represent the total number of floating-point operations during a single measured denoising stage. The count includes attention, routing, compensation, projection, feedforward networks, and sampling updates across all denoising steps.

\subsection{Evaluation Subset}
\label{sec:source206}
Due to computational constraints, we selected evaluation subsets from each benchmark by content category. HunyuanVideo used 50 long prompts from VBench, expanded by GPT. Wan2.2 used 50 image-text pairs from VBench-I2V, with 25 pairs each for the subject and background categories. For audio and video generation, H3-Base and LTX-2.5 share 50 prompts from T2AV-Compass, while FastH3 uses an expanded set that includes these 50 prompts and 50 additional prompts. SANA-WM uses 50 scenes from SANA-WM-Bench, with each scene paired with one simple camera trajectory and one difficult camera trajectory, for a total of 100 sets.

\section{Method Details}
\label{sec:source208}
\subsection{Block-Center Correlation}
\label{sec:source209}
The first level uses block-center correlation between query groups and key-value blocks to evaluate the overall response. Full blocks with scores above the first-level threshold enter the set $\mathcal{F}_{i}$ of blocks selected for exact computation, and all tokens in these blocks participate in attention calculations using their original keys and values; the remaining blocks proceed to second-level filtering. Block-center correlation is defined as:

\begin{equation}
s_{ij}^{(1)}=a_{ij}=\frac{\bar{\mathbf q}_i^\top\bar{\mathbf k}_j}{\sqrt d}.
\label{eq:source211}
\end{equation}
\subsection{Constructing the Representative Approximation Difficulty Score}
\label{sec:source212}
Block-center correlation summarizes a block's average response but does not capture variations in keys and values within the block. We measure content differences using the root-mean-square deviation of the value from the block mean:

\begin{equation}
R_j=\left(\frac{1}{n_jd_v}
\sum_{t\in\mathcal B_j}\|\mathbf v_t-\bar{\mathbf v}_j\|_2^2\right)^{1/2},\qquad \mathbf v_t,\bar{\mathbf v}_j\in\mathbb R^{d_v},
\label{eq:source214}
\end{equation}
Where $\mathbf v_t$ is the value of token t, $\bar{\mathbf v}_j$ is the block mean, $d_{v}$ is the value dimension, and $n_{j}$ is the number of tokens in the block. A larger $R_{j}$ indicates greater variability among members that the mean cannot capture.

To characterize variability of the Key along the query direction, we first compute the variance of the Key across each dimension in the block and use its diagonal statistics to estimate projection variance:

\begin{equation}
\begin{gathered}
\sigma_{j,c}^2=\frac{1}{n_j}\sum_{t\in\mathcal B_j}
(k_{t,c}-\bar k_{j,c})^2,\\[4pt]
\widehat\nu_{ij}=\frac{1}{d}\sum_{c=1}^{d}
\bar q_{i,c}^{\,2}\sigma_{j,c}^{2},
\qquad
\Delta_{ij}=\frac12\widehat\nu_{ij}.
\end{gathered}
\label{eq:source217}
\end{equation}
The projected variance estimate $\widehat{\nu}_{ij}$ describes variation in keys along the query direction that the block mean does not capture, and its calculation reuses the block's variance summary. Combining the overall response, block size, and the two types of variation above, the second-level representative approximation difficulty score is defined as:

\begin{equation}
s_{ij}^{(2)}=a_{ij}+\log(n_jR_j)+\Delta_{ij}.
\label{eq:source219}
\end{equation}
This formula combines block proxy weights, Value dispersion, and variation in keys along the query direction into a single proxy for representative approximation difficulty. Its logarithmic form lets these multiplicative factors be incorporated uniformly into the statistical threshold calculation. For each query group, calculate the mean and standard deviation of the second-level representative approximation difficulty scores only for non-Full candidate blocks, and build the threshold from these values. Blocks with scores above the threshold enter the Mixed block set $\mathcal{M}_{i}$ and proceed to token-level filtering, while the remaining blocks enter the Pure block set. The relationship between the score and the representative approximation error is described in Appendix \ref{sec:source252}.

\subsection{Raw Token Relevance}
\label{sec:source221}
For a given query group, merge the tokens from all Mixed blocks into a common set of candidates and calculate raw token relevance using the original key:

\begin{equation}
\mathcal C_i^{(3)}=\bigcup_{j\in\mathcal M_i}\mathcal B_j,
\qquad
s_{it}^{(3)}=\frac{\bar{\mathbf q}_i^\top\mathbf k_t}{\sqrt d},
\quad t\in\mathcal C_i^{(3)}.
\label{eq:source223}
\end{equation}
The third-level mean and standard deviation are computed over all token candidates in Mixed blocks. All candidates share the same query reference and threshold. Members whose raw token relevance exceeds the threshold are selected for exact computation at the third level, forming the set $\mathcal{S}_{i}^{(3)}$. Together with all members of the Full blocks, these constitute:

\begin{equation}
\mathcal E_i=
\left(\bigcup_{j\in\mathcal F_i}\mathcal B_j\right)
\cup\mathcal S_i^{(3)}.
\label{eq:source225}
\end{equation}
This comparison preserves the original response differences among the Mixed blocks. The group mean of the queries determines the shared routing result, while the final attention output is still computed separately for each original query in the group.

\subsection{Residual Interactions and Joint Normalization}
\label{sec:source227}
For query group i and key-value block j, let the number of members computed exactly and the number of remaining members be:

\begin{equation}
m_{ij}=|\mathcal E_i\cap\mathcal B_j|,
\qquad r_{ij}=n_j-m_{ij}.
\label{eq:source229}
\end{equation}
For Full blocks, $r_{ij}=0$; for Pure blocks, $r_{ij}=n_j$; and for Mixed blocks, members are either computed exactly or represented by the block summary. Residual interactions still use the original block summary to build representative terms, and after Mixed filtering, the residual subset mean is not recalculated. Let the representative logit and value be denoted as $\widetilde\ell_{ij}(\mathbf q)$ and $\widetilde{\mathbf v}_{ij}(\mathbf q)$, respectively, with $r_{ij}$ weighting the representative term. For an original query $\mathbf q$ within the query group, the output is:

\begin{equation}
\widehat{\mathbf o}_i(\mathbf q)=
\frac{
\sum_{t\in\mathcal E_i}e^{\mathbf q^\top\mathbf k_t/\sqrt d}\mathbf v_t+
\sum_{j:r_{ij}>0}r_{ij}e^{\widetilde\ell_{ij}(\mathbf q)}\widetilde{\mathbf v}_{ij}(\mathbf q)
}{
\sum_{t\in\mathcal E_i}e^{\mathbf q^\top\mathbf k_t/\sqrt d}+
\sum_{j:r_{ij}>0}r_{ij}e^{\widetilde\ell_{ij}(\mathbf q)}
}.
\label{eq:source231}
\end{equation}
Each exact term uses the original key and value of its selected token. Each representative term is weighted by the number of remaining tokens in its block. Both parts are normalized with the same denominator, so the representative terms account for the remaining contributions as the set of tokens computed exactly changes.

\section{Theoretical Basis}
\label{sec:source233}
This section examines, in turn, the feasible domain and exact-computation cost of fine-grained selection, the information limits of block-center correlation, and the ordering properties of unified comparisons across Mixed blocks.

\subsection{Feasibility Domain for Fine-Grained Selection and Exact-Computation Costs}
\label{sec:source235}
\par\noindent \textbf{Proposition 1.} Given a fixed set of queries, keys, and values, the cost of exact computation per token, and budget ceiling B, assume that block-level and fine-grained selection use the same representation and normalization rules. Also assume that fine-grained selection can reproduce the sets of tokens computed exactly and the outputs of any block-level scheme. Let the feasible sets for the two types of schemes be the block-level set and the token-level set, respectively, as defined in the following equations:

\begin{equation}
\begin{gathered}
\mathcal F_{\mathrm{block}}(B)\subseteq\mathcal F_{\mathrm{token}}(B),\\[4pt]
\min_{\mathcal E\in\mathcal F_{\mathrm{token}}(B)}\mathcal L(\mathcal E)
\le
\min_{\mathcal E\in\mathcal F_{\mathrm{block}}(B)}\mathcal L(\mathcal E),
\end{gathered}
\label{eq:source237}
\end{equation}
where $\mathcal Q_i$ is the set of queries in the i-th group, and $\mathcal{L}$ is the mean squared error of the approximate output relative to the full attention output:

\begin{equation}
\mathcal L(\mathcal E)=
\frac{1}{|\mathcal Q_i|}
\sum_{\mathbf q\in\mathcal Q_i}
\left\|\widehat{\mathbf o}_{\mathcal E}(\mathbf q)-\mathbf o(\mathbf q)\right\|_2^2.
\label{eq:source239}
\end{equation}
\par\noindent \textbf{Proof.} Take any block-level plan that satisfies the budget, and expand its selected blocks into the union of the original token indices. By assumption, this set remains feasible in the fine-grained plan, with both the exact-computation cost and the output unchanged. Therefore, every block-level plan is contained in the feasible region of the fine-grained plan. Minimizing the same error function over this larger feasible region yields the above inequality. Q.E.D.

This relationship demonstrates that allowing intra-block selection preserves feasible solutions that retain block-level computation, while also increasing flexibility in allocating the budget for exact computation across blocks.

\par\noindent \textbf{Proposition 2.} Consider a query group with one query and a key-value block of size $n=n_j\ge 3$, where exact computation costs 1 per token. Both schemes use the original block mean to represent the remaining members, weight them by their remaining quantities, and normalize them together with the exact terms. Let the minimum exact-computation costs for selecting the entire block and for selecting an arbitrary subset of tokens, respectively, when the squared output error $\mathcal L(\mathcal E)\le\varepsilon$, be the two terms in the following equations. There exist key-value configurations such that, for any $0\le\varepsilon<1/4$, the following holds:

\begin{equation}
\begin{gathered}
C_{\mathrm{block}}^\star(\varepsilon)=n,
\qquad
1\le C_{\mathrm{token}}^\star(\varepsilon)\le2,\\[4pt]
\frac{C_{\mathrm{block}}^\star(\varepsilon)}
{C_{\mathrm{token}}^\star(\varepsilon)}
\ge\frac n2.
\end{gathered}
\label{eq:source243}
\end{equation}
\par\noindent \textbf{Proof.} Let $d=d_v=1$ and $q=1$, and let the logits and values within the block be:

\begin{equation}
\begin{gathered}
(z_1,\ldots,z_n)=(\ln n,-\ln n,0,\ldots,0),\\[4pt]
(v_1,\ldots,v_n)=(1,-1,0,\ldots,0).
\end{gathered}
\label{eq:source245}
\end{equation}
Setting $k_t=z_t$ yields the above logits, and the mean of the original block satisfies $\bar k_j=\bar v_j=0$. The output of the full attention model and the output of the empty-set approximation satisfy:

\begin{equation}
\begin{gathered}
o(q)=\frac{n-n^{-1}}{n+n^{-1}+n-2}
=\frac{n^2-1}{2n^2-2n+1}>\frac12,\\[4pt]
\widehat o_{\varnothing}(q)=0,
\qquad
\mathcal L(\varnothing)>\frac14.
\end{gathered}
\label{eq:source247}
\end{equation}
Therefore, the empty set does not satisfy the given error requirements, and the full-block approach must compute all n members exactly. Let the set of tokens selected for exact computation be $\mathcal E=\{1,2\}$; assign the logits and values of the remaining members to those of the original block representative. After weighting by the remaining number $n-2$, both the numerator and denominator match those of the full attention mechanism; thus:

\begin{equation}
\mathcal L(\{1,2\})=0.
\label{eq:source249}
\end{equation}
The fine-grained approach computes at most 2 tokens exactly; because an empty set is infeasible, its exact-computation cost is at least 1, yielding the cost ratio mentioned above. Q.E.D.

This proposition compares the exact-computation costs of optimal feasible solutions, demonstrating that the gap between block-level and fine-grained decision-making can increase as the block size grows.

\subsection{Limitations of Block-Center Correlation and Criteria for Scoring Representative Approximation Difficulty}
\label{sec:source252}
\par\noindent \textbf{Proposition 3.} Even if the block-center correlation, block size, and mean value are the same, the approximation errors obtained using the block means as representatives can still differ.

\par\noindent \textbf{Proof.} Let $\bar{\mathbf q}_i$ represent a fixed query group. Let the offset of the logit within a block relative to the block-center correlation $a_{ij}$ and the value residual be denoted as:

\begin{equation}
\delta_{it}=\frac{\bar{\mathbf q}_i^\top(\mathbf k_t-\bar{\mathbf k}_j)}{\sqrt d},
\qquad \boldsymbol{\varepsilon}_t=\mathbf v_t-\bar{\mathbf v}_j,
\quad t\in\mathcal B_j.
\label{eq:source255}
\end{equation}
Definition:

\begin{equation}
\zeta_{ij}=\frac{1}{n_j}\sum_{t\in\mathcal B_j}e^{\delta_{it}},
\qquad
\boldsymbol{\xi}_{ij}=\frac{1}{n_j}\sum_{t\in\mathcal B_j}
(e^{\delta_{it}}-1)\boldsymbol{\varepsilon}_t.
\label{eq:source257}
\end{equation}
Since the sum of the Value residuals is zero, the exact denominator and numerator contributions for this block can be written as:

\begin{equation}
Z_{ij}=n_j e^{a_{ij}}\zeta_{ij},
\qquad
\mathbf N_{ij}=n_j e^{a_{ij}}
\left(\zeta_{ij}\bar{\mathbf v}_j+\boldsymbol{\xi}_{ij}\right).
\label{eq:source259}
\end{equation}
Therefore, the difference between the normalized exact output within a block and the output representing the mean is:

\begin{equation}
\mathbf o_{ij}^{\mathrm{block}}-\bar{\mathbf v}_j=\frac{\boldsymbol{\xi}_{ij}}{\zeta_{ij}}.
\label{eq:source261}
\end{equation}
When the logits within a block are not all identical, the exponential weights are uneven, and the corresponding linear mapping in the zero-mean residual space is not always zero. Thus, some zero-mean residuals make $\boldsymbol{\xi}_{ij}$ nonzero. By keeping the Key constant and scaling this residual, one can change this error while preserving the block-center correlation, the block size, and the mean of Value. Hence, these mean statistics are insufficient to determine the reliability of the approximation. Q.E.D.

In particular, $\zeta_{ij}$ depends on changes in Key along the query direction, while $\boldsymbol{\xi}_{ij}$ also depends on how these changes match Value residuals. For a single query, let $\mathbf N$ and Z denote the numerator and denominator of the full attention, and $\widehat{\mathbf N}$ and $\widehat{Z}$ their approximations; then the normalized output error satisfies:

\begin{equation}
\widehat{\mathbf o}-\mathbf o=
\frac{(\widehat{\mathbf N}-\mathbf N)-\mathbf o(\widehat Z-Z)}{\widehat Z},
\qquad \mathbf o=\frac{\mathbf N}{Z}.
\label{eq:source264}
\end{equation}
Therefore, the approximation's reliability depends on both total weight and changes in weighted content. In Appendix \ref{sec:source212}, $R_{j}$ describes the magnitude of the Value residuals, and $\Delta_{ij}$ is built from the variance of the Key in the query direction. Let $\nu_{ij}$ be the true directional variance of the centered logit. Under a second-order approximation that neglects higher-order cumulants, and by substituting the diagonal variance estimate from Appendix \ref{sec:source212} for the true directional variance, we have:

\begin{equation}
\nu_{ij}=\frac{1}{n_j}\sum_{t\in\mathcal B_j}\delta_{it}^2,
\qquad
\zeta_{ij}\approx e^{\nu_{ij}/2}\approx e^{\Delta_{ij}}.
\label{eq:source266}
\end{equation}
By combining the estimated total weight with the Value dispersion, we obtain a difficulty proxy:

\begin{equation}
\begin{gathered}
\widetilde D_{ij}=n_j e^{a_{ij}+\Delta_{ij}}R_j,\\[4pt]
\log\widetilde D_{ij}=a_{ij}+\log(n_jR_j)+\Delta_{ij}.
\end{gathered}
\label{eq:source268}
\end{equation}
This structure uses low-cost statistics to supplement information not captured by block-center correlations, and Appendix \ref{sec:source290} demonstrates its ability to distinguish levels of representative approximation difficulty.

\subsection{Sorting Properties of Uniform Comparisons Across Mixed Blocks}
\label{sec:source270}
\par\noindent \textbf{Proposition 4.} Fix the query group representative, the token candidates in all Mixed blocks, the same exact-computation cost for every candidate token, and the selection count k. Let $\mathbf z$ denote the vector of raw token relevance for each candidate under the common query reference, where $z_{t}$ is the component for token t. Define the proxy attention weight for this token as:

\begin{equation}
\begin{gathered}
\mathbf z=(z_t)_{t\in\mathcal C_i^{(3)}}\in\mathbb R^{\lvert\mathcal C_i^{(3)}\rvert},\quad z_t=s_{it}^{(3)},\\[3pt]
p_t=\frac{e^{z_t}}{\sum_{u\in\mathcal C_i^{(3)}}e^{z_u}}.
\end{gathered}
\label{eq:source272}
\end{equation}
A set formed by selecting the k members with the highest scores globally ensures that the retained proxy attention mass is no less than that of any other set of the same size:

\begin{equation}
\begin{gathered}
\mathcal S^\star=\operatorname{TopK}(\mathbf z,k)
\quad\Longrightarrow\\
\sum_{t\in\mathcal S^\star}p_t
\ge \sum_{t\in\mathcal S}p_t,
\\
\mathcal S\subseteq\mathcal C_i^{(3)},\quad|\mathcal S|=k.
\end{gathered}
\label{eq:source274}
\end{equation}
\par\noindent \textbf{Proof.} Since all candidates share the same normalized denominator and the exponential function is strictly monotonic, the proxy weights preserve the ordering by raw token relevance. If a set keeps a lower-scoring member while omitting a higher-scoring one, swapping them will not reduce the retained attention mass. Repeating this swap yields the set with the highest global score. Q.E.D.

Sol\textsuperscript{3} uses a common threshold for all tokens in Mixed blocks, so it retains the globally highest-scoring tokens for the actual selection size. Uniform cross-block comparison lets the budget for exact computation be allocated across all candidates, maximizing the retained proxy attention mass under a common query reference.

\section{Efficient Kernel Implementation}
\label{sec:source277}
Sol\textsuperscript{3}'s hierarchical routing reduces the number of attention interactions requiring exact computation. However, when the same query group selects a small number of tokens from each of multiple Mixed blocks, the original block structure still splits these interactions into multiple small tasks. The denoising layer still needs to prepare attention inputs and update hidden states for each video token. If adjacent operators are executed separately, intermediate features meant only for the next step must first be written back to GPU global memory and then read by the next operator. To address these overheads, we customized a set of Triton kernels for Sol\textsuperscript{3}. We store the token indices selected from each query group consecutively in a single row of a compact index (\ref{sec:source278}), allowing the exact-attention kernel to process these interactions in batches across blocks. The merged kernel weights the representative terms by the number of remaining members in each block and normalizes them together with the exact term (\ref{sec:source281}). During input preparation and residual updates, we also merge adjacent operators, removing the need to write intermediate features back to and re-read them from GPU global memory (\ref{sec:source1503}).

\subsection{Compact Representation of the Exact Computation Set for a Query Group}
\label{sec:source278}
For query group $Q_{i}$, all Full blocks selected by first-level routing are computed exactly. The second level splits the remaining blocks into Mixed and Pure blocks, with Pure blocks using representative approximation. For Mixed blocks, the third level applies shared threshold filtering to all candidate tokens in the group and records selected members in a bitmap. Because the block state and member bitmap are fixed for each query group, and queries within a group share the same set $\mathcal{E}_{i}$ of tokens for exact computation, they can be represented by a single row in the Compressed Sparse Row (CSR) index. We determine each block's write positions in the row from the prefix sum of selection counts, and store the original IDs of selected tokens in sequential order, enabling the exact-attention kernel to process these members in batches across the original blocks.

\subsection{Route Score Reuse, Cross-Block Exact Computation, and Representation Merging}
\label{sec:source281}
When constructing block summaries, the routing kernel computes block means for Query, Key, and Value, along with the dispersion statistics needed for routing, in a single read operation. For Mixed blocks, the same dot product estimates the threshold and filters members. Algorithm \ref{alg:csr} stores these scores while it accumulates block statistics, then aggregates statistics from all Mixed blocks in the same query group to obtain a shared threshold. The kernel then directly compares the stored scores with the threshold to generate the membership bitmap, avoiding a second round of dot product calculations.

\begin{algorithm}[!htbp]
\caption{Mixed-Block Score Reuse and CSR Construction}
\label{alg:csr}
\small
{\raggedright
\noindent \textbf{Require: } Query-group means $\{\bar{\mathbf q}_i\}_{i=0}^{G-1},\ K,$ key blocks $\{\mathcal B_j\}$ of sizes $\{n_j\},$ Full and Mixed block sets $\{\mathcal F_i,\mathcal M_i\},$ threshold parameter $\beta_3.$\par
\noindent \textbf{Ensure: } CSR indices $(\mathrm{ptr},\mathrm{idx})$ for $\{\mathcal E_i\},$ remaining counts $\{r_{ij}\}.$\par
}
\begin{tabular}{@{}r@{\enspace}>{\raggedright\arraybackslash}p{\dimexpr\linewidth-2em\relax}@{}}
1: & $\textbf{for each }(i,j)\text{ with }j\in\mathcal M_i\textbf{ in parallel do}$ \\
2: & \mbox{$\quad \mathbf s^{(3)}_{i,\mathcal B_j}\gets K_{\mathcal B_j}\bar{\mathbf q}_i/\sqrt d$}\hspace{.44em}\mbox{//\enspace\textcolor[HTML]{808080}{save scores for token selection}} \\
3: & $\quad \mu_{ij}\gets\operatorname{mean}(\mathbf s^{(3)}_{i,\mathcal B_j}),\allowbreak\qquad M_{2,ij}\gets\sum_{t\in\mathcal B_j}(s^{(3)}_{it}-\mu_{ij})^2$ \\
4: & $\textbf{end for}$ \\
5: & $\textbf{for each }\text{query group }i\text{ with }\mathcal M_i\ne\varnothing\textbf{ in parallel do}$ \\
6: & $\quad (\mu_i^{(3)},\sigma_i^{(3)})\gets\operatorname{PooledMeanStd}\!\bigl(\{(n_j,\mu_{ij},M_{2,ij})\}_{j\in\mathcal M_i}\bigr)$ \\
7: & \mbox{$\quad \tau_i^{(3)}\gets\mu_i^{(3)}+\beta_3\sigma_i^{(3)}$}\hspace{.44em}\mbox{//\enspace\textcolor[HTML]{808080}{one threshold across Mixed blocks}} \\
8: & $\quad \mathrm{bits}_{ij}\gets\operatorname{PackBits}(\mathbf s^{(3)}_{i,\mathcal B_j}>\tau_i^{(3)}),\allowbreak\qquad j\in\mathcal M_i$ \\
9: & $\textbf{end for}$ \\
10: & $m_{ij}\gets\begin{cases}n_j & j\in\mathcal F_i\\ \operatorname{popcount}(\mathrm{bits}_{ij}) & j\in\mathcal M_i\\ 0 & \text{otherwise}\end{cases},\allowbreak\qquad r_{ij}\gets n_j-m_{ij}$ \\
11: & $\mathrm{ptr}_i\gets\sum_{i'<i}\sum_j m_{i'j},\allowbreak\qquad 0\le i\le G$ \\
12: & $\textbf{for each }(i,j)\text{ with }m_{ij}>0\textbf{ in parallel do}$ \\
13: & \mbox{$\quad o\gets\mathrm{ptr}_i+\sum_{j'<j}m_{ij'}$}\hspace{.44em}\mbox{//\enspace\textcolor[HTML]{808080}{first write position in idx}} \\
14: & $\quad \mathrm{idx}[o:o+m_{ij}]\gets\begin{cases}\mathcal B_j & j\in\mathcal F_i\\ \operatorname{MaskedSelect}(\mathcal B_j,\mathrm{bits}_{ij}) & j\in\mathcal M_i\end{cases}$ \\
15: & $\textbf{end for}$ \\
16: & $\textbf{return }\mathrm{ptr},\mathrm{idx},\{r_{ij}\}$ \\
\end{tabular}
\end{algorithm}
Algorithm \ref{alg:aggregation} weights each representative term by the number of tokens in each key-value block that did not participate in the exact computation. If $m_{ij}$ of the $n_{j}$ tokens in block $\mathcal{B}_{j}$ participate in the exact computation, then the block's numerical weight is $r_{ij}=n_j-m_{ij}$. For blocks with $r_{ij} > 0$, the merging kernel adds $\log r_{ij}$ to the block's representative logit, then uses the logarithmic normalization term $L^{\mathrm{ex}}$ from the exact-attention kernel to merge the exact computation term with the representative term according to the formula in Appendix \ref{sec:source227}.

\begin{algorithm}[!htbp]
\caption{Indexed Exact Attention and Representative Aggregation}
\label{alg:aggregation}
\small
{\raggedright
\noindent \textbf{Require: }$Q,K,V,$ CSR indices $(\mathrm{ptr},\mathrm{idx}),$ remaining counts $\{r_{ij}\},$ block summaries for $\widetilde\ell_{ij}$ and $\widetilde{\mathbf v}_{ij}.$\par
\noindent \textbf{Ensure: } Attention output $\widehat O$ in query order.\par
}
\begin{tabular}{@{}r@{\enspace}>{\raggedright\arraybackslash}p{\dimexpr\linewidth-2em\relax}@{}}
1: & $\textbf{for each }\text{query }\mathbf q\in Q_i\text{ across all groups }i\textbf{ in parallel do}$ \\
2: & $\quad \mathcal E_i\gets\mathrm{idx}[\mathrm{ptr}_i:\mathrm{ptr}_{i+1}]$ \\
3: & $\quad (\mathbf o^{\mathrm{ex}},L^{\mathrm{ex}})\gets\operatorname{ExactWithLSE}(\mathbf q,K,V,\mathcal E_i)$ \\
4: & \mbox{$\quad (u,z,\mathbf a)\gets(-\infty,0,\mathbf 0)$}\hspace{.44em}\mbox{//\enspace\textcolor[HTML]{808080}{online softmax over representatives}} \\
5: & $\quad \textbf{for each }j\text{ with }r_{ij}>0\textbf{ do}$ \\
6: & \mbox{$\qquad s\gets\widetilde\ell_{ij}(\mathbf q)+\log r_{ij},\enspace u'\gets\max(u,s)$}\hspace{.44em}\mbox{//\enspace\textcolor[HTML]{808080}{weight by the remaining token count}} \\
7: & $\qquad \alpha\gets e^{u-u'},\allowbreak\qquad p\gets e^{s-u'}$ \\
8: & $\qquad (\mathbf a,z,u)\gets\bigl(\alpha\mathbf a+p\,\widetilde{\mathbf v}_{ij}(\mathbf q),\ \alpha z+p,\ u'\bigr)$ \\
9: & $\quad \textbf{end for}$ \\
10: & $\quad h\gets\max(L^{\mathrm{ex}},u),\allowbreak\qquad w_{\mathrm{ex}}\gets e^{L^{\mathrm{ex}}-h},\allowbreak\qquad w_{\mathrm{rep}}\gets e^{u-h}$ \\
11: & \mbox{$\quad \widehat{\mathbf o}_i(\mathbf q)\gets\frac{w_{\mathrm{ex}}\mathbf o^{\mathrm{ex}}+w_{\mathrm{rep}}\mathbf a}{w_{\mathrm{ex}}+w_{\mathrm{rep}}z}$}\hspace{.44em}\mbox{//\enspace\textcolor[HTML]{808080}{merge using the exact log normalizer}} \\
12: & $\textbf{end for}$ \\
13: & $\textbf{return }\operatorname{Stack}_i(\widehat O_i)$ \\
\end{tabular}
\end{algorithm}
\subsection{Operator Fusion in Attention Input Preparation and Residual Updates}
\label{sec:source1503}
For each video token, Sol\textsuperscript{3} applies a conditional scale and shift to its normalized features in a single Triton kernel. This avoids writing the scaled features to GPU global memory and reading them back before the shift. The modulated features are then projected to yield Q, K, and V. A fused Q/K kernel applies rotary position embedding (RoPE) immediately after the final scaling step in Q/K normalization, so the scaled Q and K need not be written to GPU global memory and read back for rotation. The rotated key contributes to the block mean used for first-level routing, while the token's own query is used to compute its attention output. After the output projection, a single kernel gates the projected output and adds it to the token's hidden state. This fusion avoids writing the gated output to GPU global memory and reading it back for residual addition.

\section{Further Analysis of GHR}
\label{sec:source284}
\subsection{The Phenomenon of Minority Dominance}
\label{sec:source285}
We compare inter-block and intra-block attention distributions across models. A few blocks account for most of the attention, while some low-response blocks still contain a few high-response tokens.

\begin{figure*}[!tp]
\centering
\includegraphics[width=\linewidth]{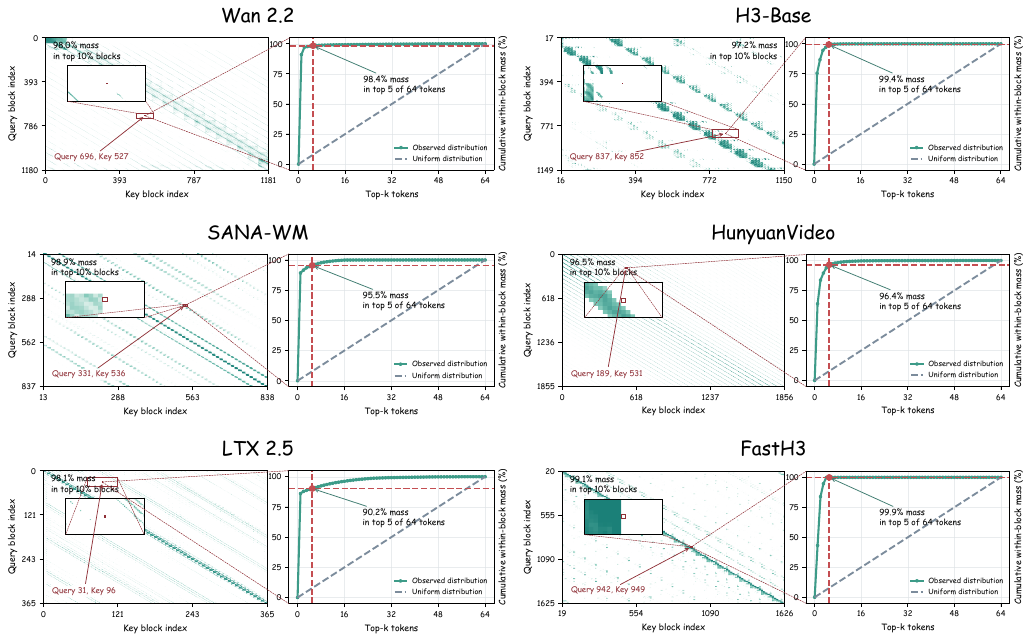}
\caption{\textbf{Additional examples of attention concentration across and within blocks.}}
\label{fig:F1}
\end{figure*}
\subsection{Gaussian Shape and Selection Ratio Calibration}
\label{sec:source287}
\begin{figure*}[!tp]
\centering
\includegraphics[width=\linewidth]{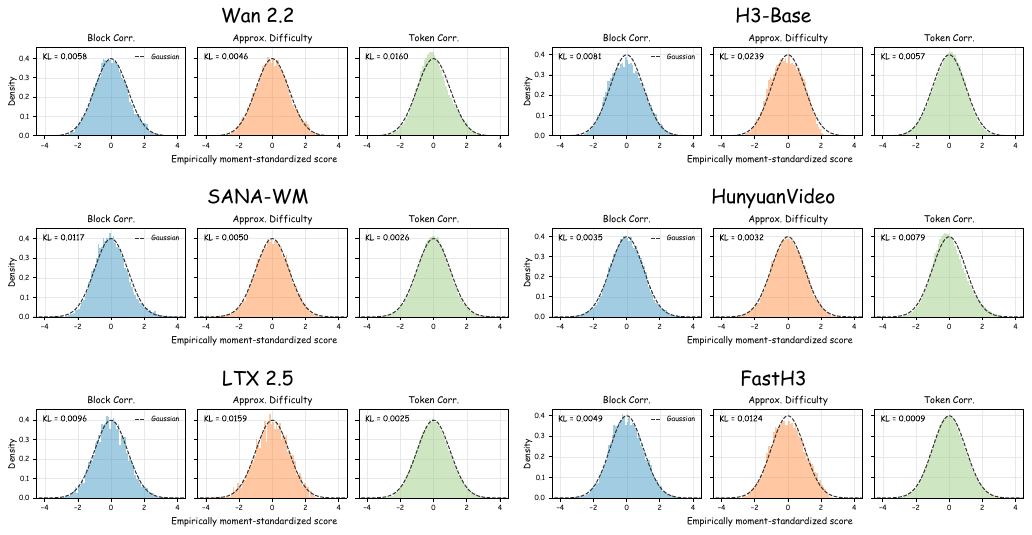}
\caption{\textbf{Comparison of the distribution of three-level routing scores for additional models with a Gaussian reference.}}
\label{fig:F2}
\end{figure*}

\begin{figure*}[!tp]
\centering
\includegraphics[width=\linewidth]{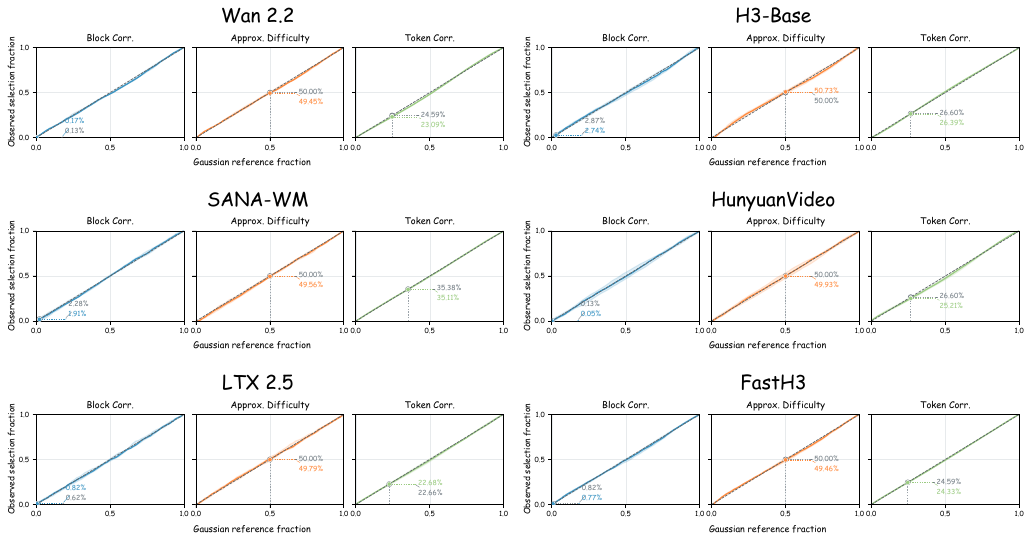}
\caption{\textbf{Actual selection ratios versus Gaussian reference ratios for additional models.}}
\label{fig:F3}
\end{figure*}

Figure \ref{fig:F2} shows the distribution of three-level routing scores for the six models. The scores at each level are approximately Gaussian. In Figure \ref{fig:F3}, we compare the actual selection rates at different filtering thresholds with the corresponding Gaussian reference rates, and find that the two are generally close.

\section{Further Analysis of HRS}
\label{sec:source289}
\subsection{Representative Approximation Difficulty}
\label{sec:source290}
We present pairs of blocks from different models with similar block-center correlations but different approximation errors, and we show the relationship between the representative approximation difficulty score and approximation error. Compared with block-center correlation, the full score better reflects the actual variation in approximation error.

\begin{figure*}[!tp]
\centering
\includegraphics[width=\linewidth]{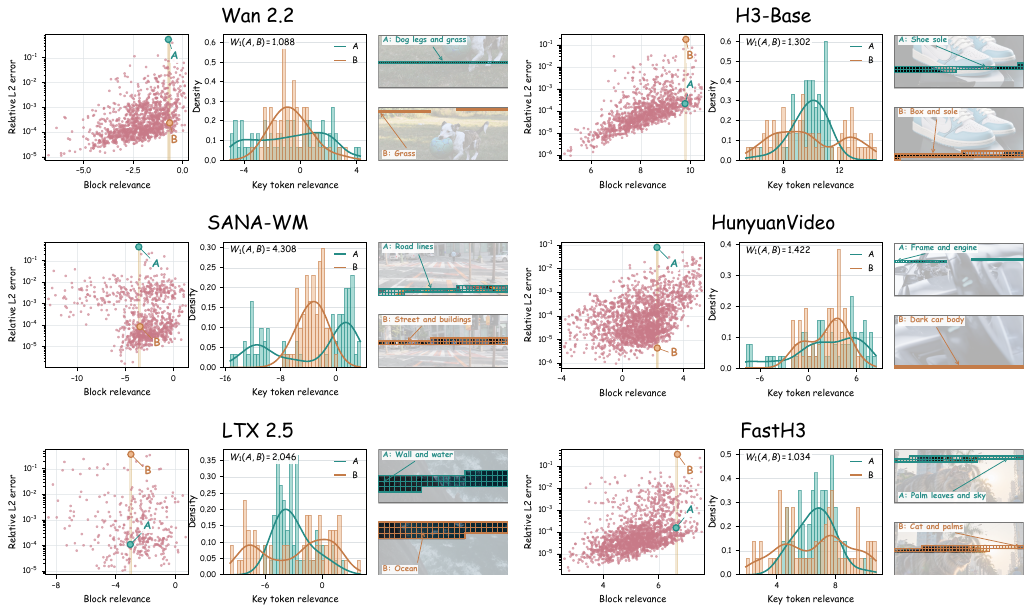}
\caption{\textbf{Additional examples of block pairs in additional models that have similar block-center correlations but represent different levels of approximation error.}}
\label{fig:G1}
\end{figure*}
\begin{figure*}[!tp]
\centering
\includegraphics[width=\linewidth]{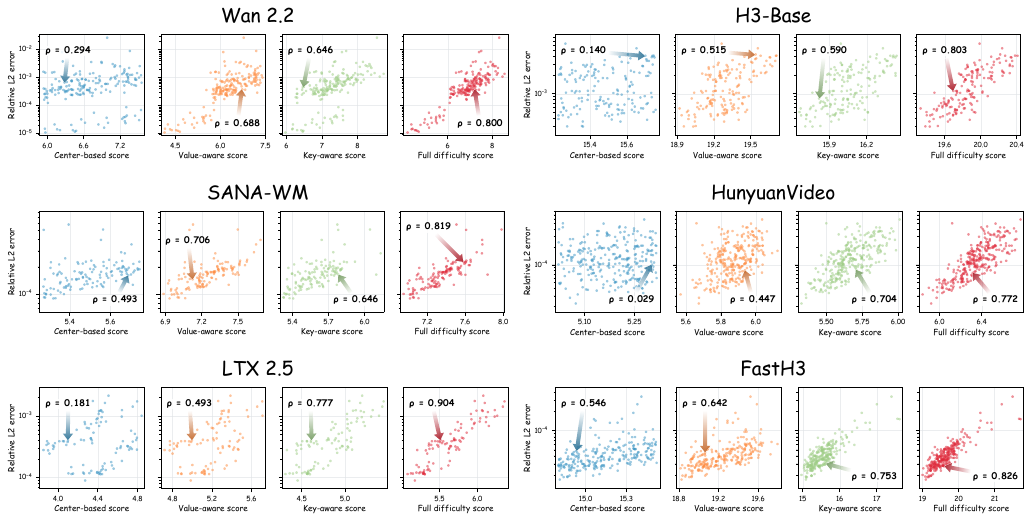}
\caption{\textbf{Visualizations of representative approximation difficulty and approximation error in additional models.}}
\label{fig:G2}
\end{figure*}
\subsection{Selection of Members Across Mixed Blocks}
\label{sec:source292}
We present results on cross-block token selection across multiple models and compare filtering methods under the same budget for exact computation. Uniform filtering across Mixed blocks preserves more key interactions selected by the Mass oracle than block-wide filtering and independent block-by-block filtering. Compared with independent block-by-block filtering, uniform filtering preserves a larger fraction of the attention mass and shows lower relative $L_{2}$ error in attention outputs.

\begin{figure*}[!tp]
\centering
\includegraphics[width=\linewidth]{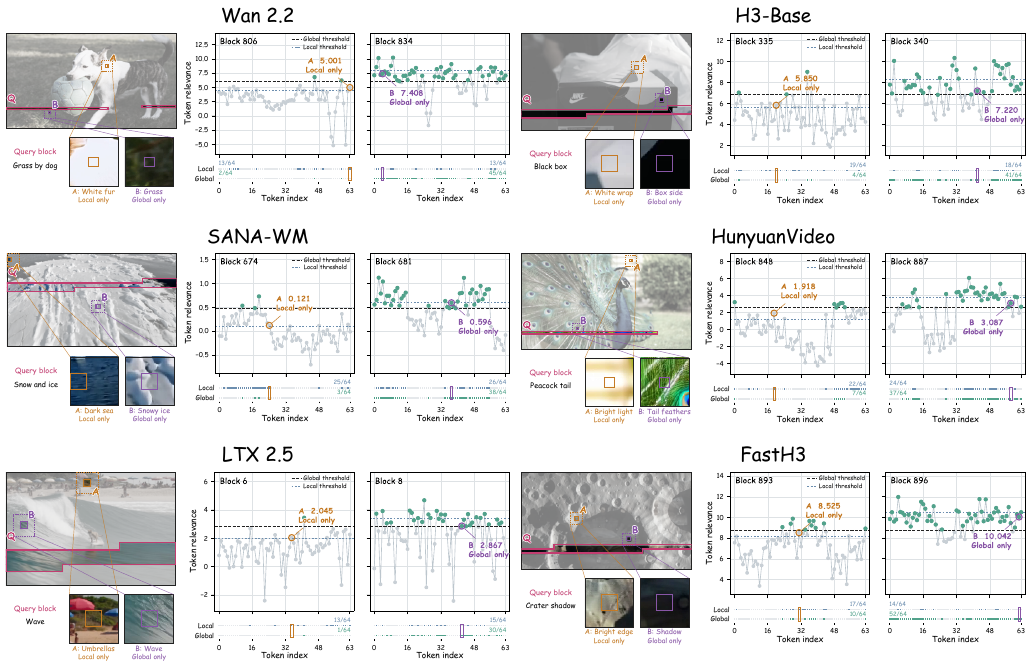}
\caption{\textbf{Token selection results for block-by-block filtering and unified filtering across Mixed blocks on additional models.}}
\label{fig:G3}
\end{figure*}
\begin{figure*}[!tp]
\centering
\includegraphics[width=\linewidth]{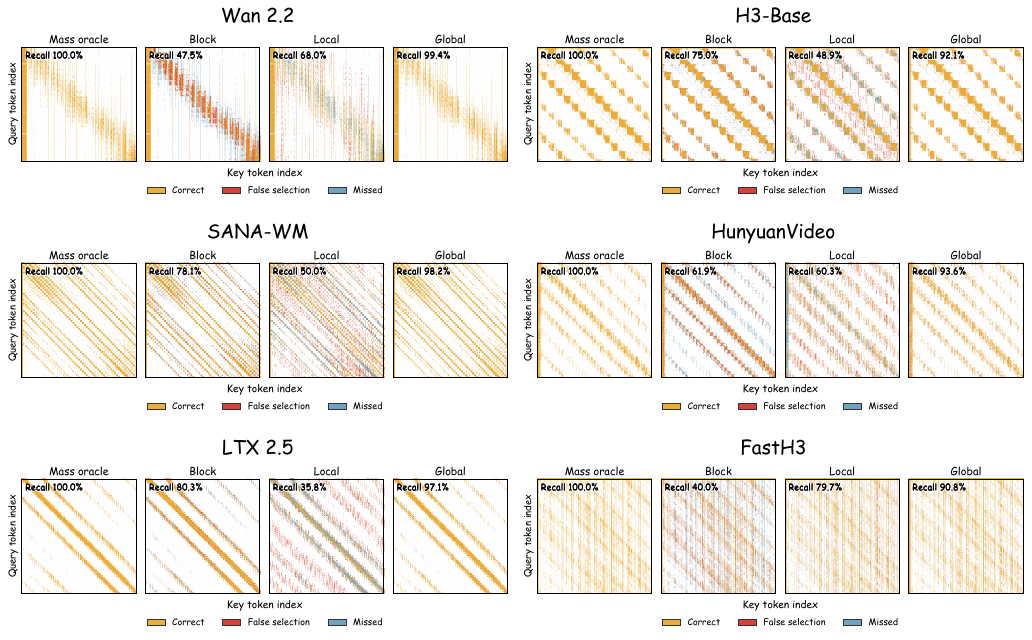}
\caption{\textbf{Visualization of how different filtering methods preserve key interactions across multiple models.}}
\label{fig:G4}
\end{figure*}
\begin{figure*}[!tp]
\centering
\includegraphics[width=\linewidth]{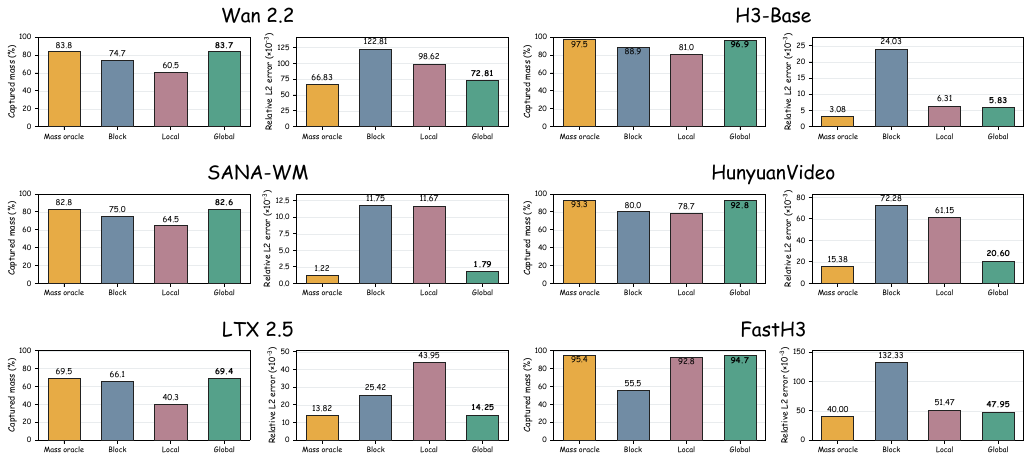}
\caption{\textbf{Fraction of attention mass retained and relative $L_{2}$ output error for different filtering methods across multiple models.}}
\label{fig:G5}
\end{figure*}
\section{Additional Visual Comparisons}
\label{sec:source294}
\begin{figure*}[t!]
\centering
\includegraphics[width=\linewidth]{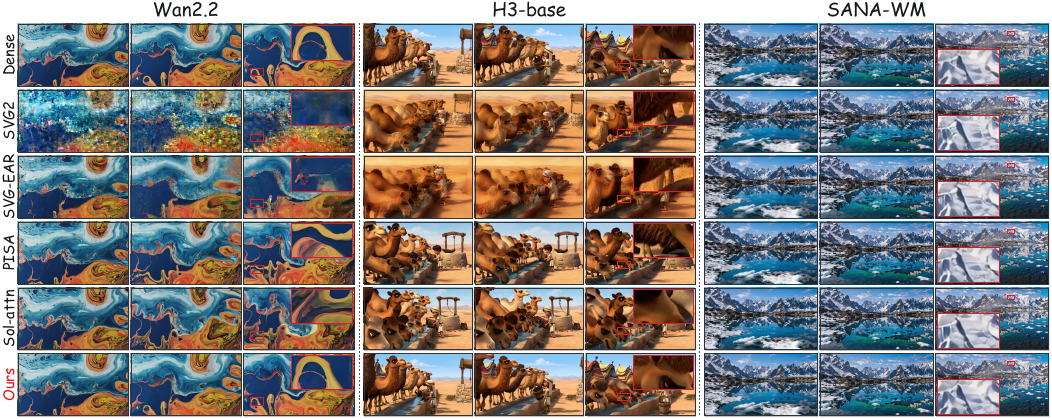}
\caption{\textbf{Additional visual comparisons between \textbf{Sol\textsuperscript{3}} and existing methods.}}
\label{fig:H1}
\end{figure*}
\begin{figure*}[t!]
\centering
\includegraphics[width=\linewidth]{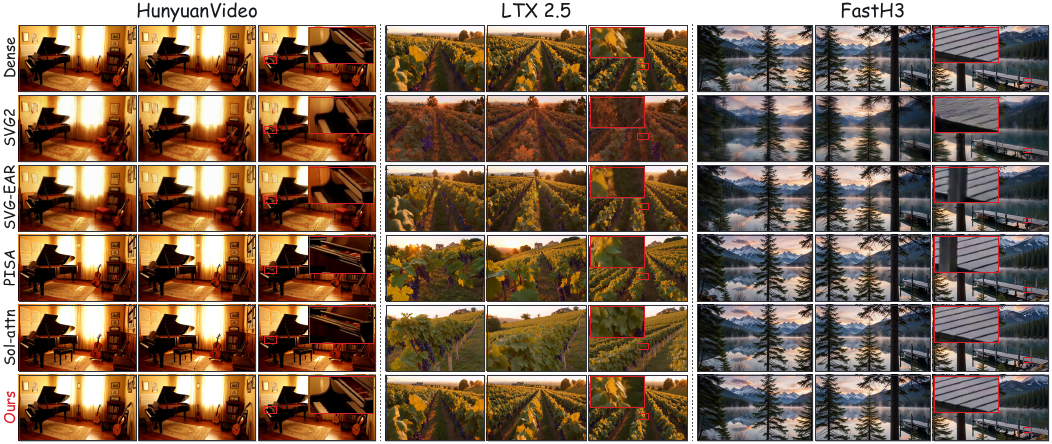}
\caption{\textbf{Additional visual comparisons between \textbf{Sol\textsuperscript{3}} and existing methods.}}
\label{fig:H2}
\end{figure*}
\begin{figure*}[t!]
\centering
\includegraphics[width=\linewidth]{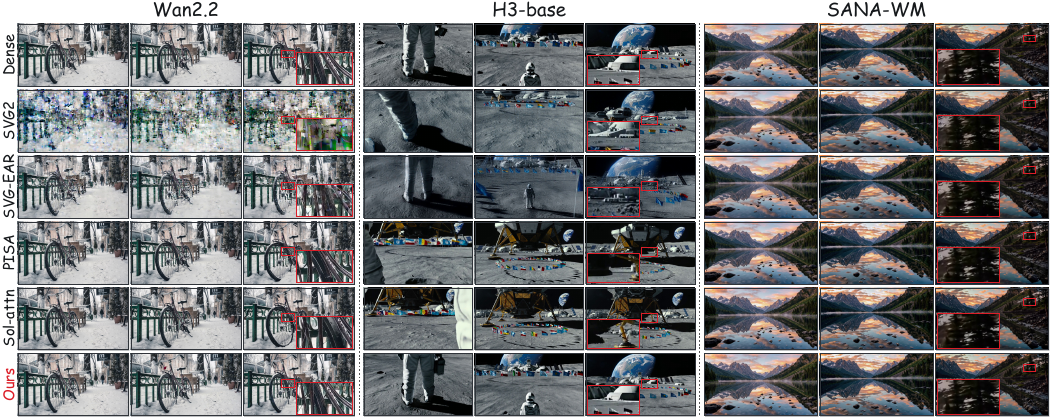}
\caption{\textbf{Additional visual comparisons between \textbf{Sol\textsuperscript{3}} and existing methods.}}
\label{fig:H3}
\end{figure*}
\begin{figure*}[t!]
\centering
\includegraphics[width=\linewidth]{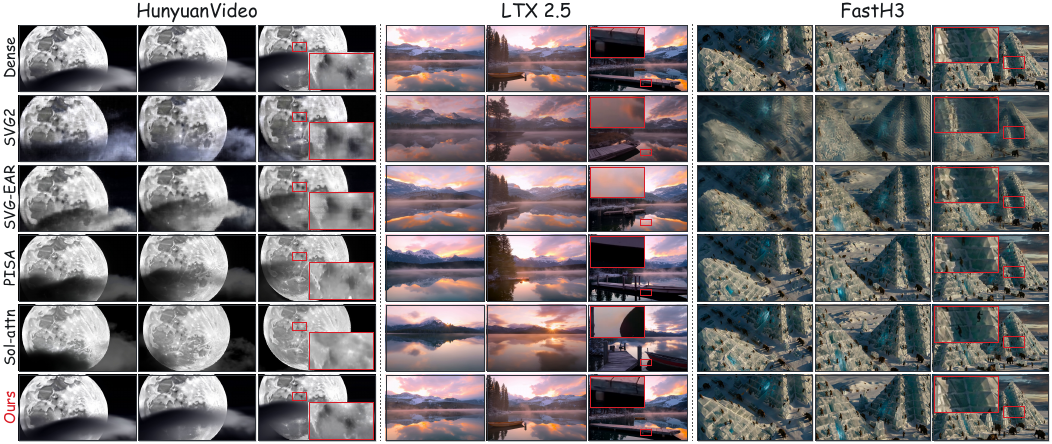}
\caption{\textbf{Additional visual comparisons between \textbf{Sol\textsuperscript{3}} and existing methods.}}
\label{fig:H4}
\end{figure*}
This section presents additional visual comparisons of Sol\textsuperscript{3} and existing sparse attention methods on Wan2.2, H3-Base, SANA-WM, HunyuanVideo, LTX-2.5, and FastH3, further demonstrating Sol\textsuperscript{3}'s ability to preserve structure and detail. The results are shown in Figures \ref{fig:H1}--\ref{fig:H4}.

\end{document}